\documentclass{article}
\usepackage{ijcai24}

\usepackage{times}
\usepackage{soul}
\usepackage{url}
\usepackage[hidelinks]{hyperref}
\usepackage[utf8]{inputenc}
\usepackage[small]{caption}
\usepackage{graphicx}
\usepackage{amsmath}
\usepackage{amsthm}
\usepackage{booktabs}
\usepackage{xcolor}
\usepackage[linesnumbered,ruled]{algorithm2e}

\SetCommentSty{mycommfont}
\usepackage{amssymb}
\usepackage[switch]{lineno}

\usepackage{subcaption}
\usepackage{amssymb}
\usepackage{mathrsfs}
\usepackage{multicol}
\usepackage{multirow}
\usepackage{xspace}
\usepackage{float}
\usepackage{makecell}
\usepackage{tabularx}

\let\oldhat\hat

\renewcommand{\hat}[1]{\oldhat{\mathbf{#1}}}

\newcommand{\etal}{\emph{et al.}}

\title{Position Debiasing Fine-Tuning for Causal Perception in Long-Term Dialogue}

\author{
Shixuan Fan$^{1,2}$\and
Wei Wei\thanks{Corresponding authors.}$^{1,2}$\and
Wendi Li$^{1,2}$\and
Xian-Ling Mao$^3$\and
Wenfeng Xie$^{2,4}$\And
\\ Dangyang Chen$^{\ast 2,4}$
\affiliations
$^1$Cognitive Computing and Intelligent Information Processing (CCIIP) Laboratory, School of Computer Science and Technology, Huazhong University of Science and Technology\\
$^2$Joint Laboratory of HUST and Pingan Property \& Casualty Research (HPL)\\
$^3$Department of Computer Science and Technology, Beijing Institute of Technology\\
$^4$Ping An Property \& Casualty Insurance Company of China, Ltd\\ 
\emails
\{fanshixuan, weiw, wendili\}@hust.edu.cn,
maoxl@bit.edu.cn,
julian\_wind@163.com,
chendangyang273@pingan.com.cn
}

\begin{document}

\maketitle

\begin{abstract}
     The core of the dialogue system is to generate relevant, informative, and human-like responses based on extensive dialogue history. Recently, dialogue generation domain has seen mainstream adoption of large language models (LLMs), due to its powerful capability in generating utterances. However, there is a natural deficiency for such models, that is, inherent position bias, which may lead them to pay more attention to the nearby utterances instead of causally relevant ones, resulting in generating irrelevant and generic responses in long-term dialogue. To alleviate such problem, in this paper, we propose a novel method, named Causal Perception long-term Dialogue framework (CPD), which employs perturbation-based causal variable discovery method to extract casually relevant utterances from the dialogue history and enhances model causal perception during fine-tuning. Specifically, a local-position awareness method is proposed in CPD for inter-sentence position correlation elimination, which helps models extract causally relevant utterances based on perturbations. Then, a casual-perception fine-tuning strategy is also proposed, to enhance the capability of discovering the causal invariant factors, by differently perturbing causally relevant and non-casually relevant ones for response generation. Experimental results on two datasets prove that our proposed method can effectively alleviate the position bias for multiple LLMs and achieve significant progress compared with existing baselines.
\end{abstract}

\section{Introduction}
\begin{figure}[!t]
    \centering
    \includegraphics[width=0.8\linewidth]{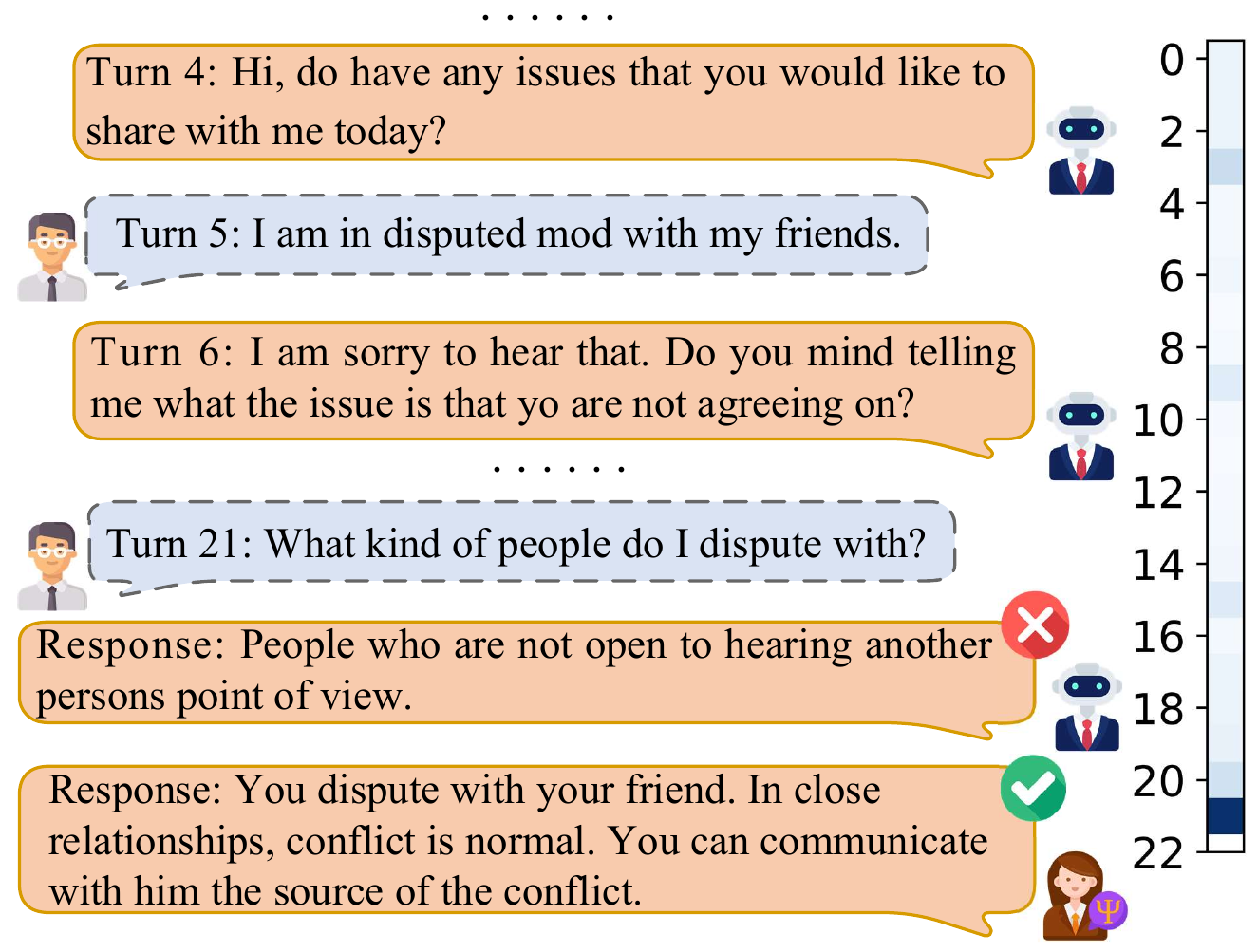}
\caption{Position bias in large language models (Llama2-7B-chat). Dotted boxes mark relevant utterances. The strip on the right shows the average attention of each turn in the dialogue. Darker colors represent higher attention values.}
\label{fig:intro}
\end{figure}

The design of the dialogue systems aspires to generate consistent, controllable, and diverse responses based on dialogue history \cite{liu2022improving,lu2023miracle}. Initial researches on dialogue systems are often constrained by the length of the input window, with a common assumption that responses are primarily influenced by the last one or a few last turns of the dialogue history \cite{wei2021target,liu2022incorporating}. Recently, with the development of large language models (LLMs), researchers propose long-term dialogue tasks and explore the use of rich information in dialogue history for response generation \cite{xu2022long}.

Despite the impressive achievements of LLMs in open-domain dialogues, capturing causal relationships within extensive dialogue history remains challenging \cite{feng2023less}. Some researchers identify that LLMs suffer from severe position bias, focusing only on the final context and disregarding historical information \cite{liu2023lost,liu2023instruction}. This bias leads models to concentrate on spurious position correlations instead of causally relevant utterances, resulting in irrelevant and generic response generation. As shown in Figure \ref{fig:intro}, Llama2's attention in the last turn is significantly higher than in other turns, neglecting the relevance utterance in the $5$-th turn and resulting in the generation of context-irrelevant and generic responses. To address the above problems, some studies aim to compress long-term dialogue history using retrieval-based \cite{feng2023less} or summary-based \cite{wang2023recursively,DBLP:conf/acl/LeeHPP023} methods. Although these works achieve certain results, none substantially improve the ability of LLMs to perceive genuinely causally relevant utterances. 

To eliminate position bias and enhance the perception of causal relationships in long-term dialogue, two significant challenges must be addressed: 
(1) Lack of large-scale dialogue datasets with causally relevant utterance annotations. Perturbation-based causal variable discovery methods assume that models can effectively utilize all input variables. They consider the difference in potential outcomes when binary intervention (presence or exclusion) acts on the variable as the treatment effect. However, the inherent position bias in LLMs hinders their ability to fully leverage dialogue history, making perturbation-based methods unable to be directly used to extract causally relevant utterances. 
(2) Traditional text generation loss is insufficient in guiding models to eliminate spurious correlations between relevant utterances and position distributions. Models tend to learn imbalances in the distribution of relevant utterance positions in datasets. Existing position debiasing methods primarily disrupt positions,  posing challenges in maintaining the consistency of semantic structures when applied to dialogue scenarios.

To address the aforementioned challenges, we propose a model-free Causal Perception long-term Dialogue framework, named CPD. 
Firstly, we extract relevant utterances for two widely used long-term dialogue datasets using causal perturbation. Analyzing the impact of position bias on the model's causal perception, we introduce a local-position awareness method to mitigate this bias by eliminating inter-sentence position information. Through sentence-level perturbations on dialogues, we regard the change in the perplexity \cite{1995From} before and after the perturbation as the treatment effect of the perturbed utterance. 
Validation of sentence-level conditional independence in sets of causally relevant utterances (refer to Appendix \ref{app:independent}), followed by the application of clustering algorithms to categorize dialogue history into causally relevant or irrelevant parts. 
Secondly, inspired by invariance learning \cite{chang2020invariant}, the core of preventing models from fitting position bias is to direct models to concentrate on invariant causal variables. We argue that responses and their corresponding causally relevant utterances should exhibit consistency under perturbations. Specifically, we conduct sentence-level perturbations separately on causally relevant and non-causally relevant parts, encouraging the model's generation to be either consistent or inconsistent with gold responses based on whether the causal variables are perturbed or not. A positional difference sampling strategy, based on the probability of relevant utterance positions, is employed to address the imbalance of relevant utterance positions without compromising the integrity of the dialogue structure.

The contributions of this paper are listed as follows:
\begin{itemize}
    \item We propose a relevant utterance extraction method based on causal perturbation, which can avoid the interference caused by the position bias of LLMs through local-position awareness.
    \item We propose a causal perception fine-tuning strategy that effectively alleviates the model's position bias and enhances its ability to perceive correlations in dialogues.
    \item Experimental results on two benchmark datasets demonstrate that our proposed method can consistently outperform the state-of-the-art baselines in terms of objective and subjective evaluation metrics.
\end{itemize}
\section{Related Work}

\subsection{Position Bias and Long-Term Dialogue}
Position bias in language models has undergone extensive examination. Ko \etal \shortcite{DBLP:conf/emnlp/KoLKKK20} observed that language models introduced position bias due to the imbalance position distribution of relevant sentences in datasets, relying on fixed-position utterances instead of real causal correlation during inference. Wang \etal \shortcite{DBLP:conf/iclr/WangSLJYLS21} and Liu \etal \shortcite{liu2023lost} respectively confirmed the existence of position bias in a variety of pre-trained models utilizing different position embeddings and a variety of LLMs of different sizes. 

Existing position debiasing methods primarily involved disrupting the position information in datasets and constraining the consistency of the original model during the fine-tuning process. Ko \etal \shortcite{DBLP:conf/emnlp/KoLKKK20} and RPP \cite{amor2023technical} applied random position perturbations at the document and word level separately. Some other works divide context into short segments \cite{ivgi2023efficient,li2023split}, or reorder input context based on relevance \cite{peysakhovich2023attention,he2023never}. However, dialogues exhibit temporal structure, and position perturbation can destroy dialogue structures, leading to a substantial decrease in comprehension. ZOE \cite{liu2024zero} required the model to fit both task labels and original model output, alleviating the impact of position imbalance by enforcing consistency between the fine-tuned model and the original model. Unfortunately, in dialogue scenarios, the inherent position bias of the model aligns with the position imbalance in the data, causing the method to fail to a certain extent. 

To explore the model's ability to comprehend rich semantics in long-term dialogue history, Xu \etal ~\shortcite{DBLP:conf/acl/XuSW22} constructed a multi-turn long-term dialogue dataset. Existing works primarily focused on enabling LLMs to acquire long-term dialogue awareness by compressing dialogue history to alleviate position bias. RSM \cite{wang2023recursively} and Lee \etal \shortcite{DBLP:conf/acl/LeeHPP023} employed LLMs to summarize dialogue history as an external memory pool, enhancing LLMs with long-term memory capabilities. CONSTRAIN \cite{feng2023less} assumed that only two sentences in the dialogue history were relevant to the response, utilizing a trained language model to retrieve relevant utterances. However, these methods suffer from dual challenges of information loss during compression and a lack of substantial improvement in the model's inherent ability to comprehend extended contexts.

\subsection{Causal Inference in NLP}
Causal inference is a method used to determine whether correlations in data truly reflect cause-and-effect relationships \cite{pearl2016causal,alaa2019validating}. Works on causal inference in NLP mainly focused on word granular detoxification in pre-trained language models. For instance, Abraham \etal \shortcite{abraham2022cebab} and Madhavan \etal \shortcite{madhavan2023cfl} both engaged in word-level language model detoxification, employing average treatment effect and counterfactual enhancement. Wang \etal \shortcite{wang2023recursively} is similar to our work. They annotate a small test set of causally relevant utterances, named CGDIALOG, and discovered that the causal perception ability of language models is deficient, meaning that language models exhibit similar outcomes to perturbations of causally relevant or non-causally relevant utterances. They simply attributed this problem to overfitting. In our investigation, we delved further and identified position bias as the root cause of this phenomenon.

\section{Methodology}

\begin{figure}[t]
  \centering
  \includegraphics[width=0.6\columnwidth]{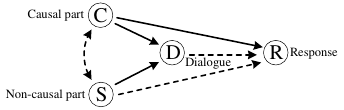}
  \caption{Causal view of response generation, where the solid line represents the causal relationship between two variables, and the dotted line represents the probabilistic dependencies.}
  \label{fig:causal}
\end{figure}

\subsection{Causal View for Response Generation}

\noindent \textbf{Structure Causal Model (SCM)}~\cite{DBLP:journals/ijon/Shanmugam01} is a method of describing causal correlations among multiple variables. In this paper, we describe the causal structure in dialogue as a causal graph composed of four variables: input dialogue $D$, response $R$, causal part $C$, and non-causal part $S$. Figure \ref{fig:causal} illustrates the SCM of dialogue. 

Dialogue $D$ consists of a causal part $C$ and a non-causal part $S$ of the response ($C\rightarrow D\leftarrow S$), where the causal part $C$ is the minimum set of corresponding endogenous cause utterances leading to the response $R$ ($C\rightarrow R$). An outstanding language model should exhibit sensitivity to causal part $C$ while maintaining robustness to non-causal part $S$. Unfortunately, due to the probabilistic correlation between the causal part $C$ and the non-causal part $S$ ($C \dashleftarrow \dashrightarrow S$), including the imbalanced distribution of causal part utterances across positions (see in Appendix \ref{app:causal_static}), often leads models to overfit position information, fixate on utterances at nearby positions, and lose the ability to perceive causality ($S \dashrightarrow R$).

\noindent \textbf{Treatment Effect (TE)}~\cite{austin2011introduction} quantifies the impact of a variable on the target outcome. In practice, TE is often assessed through the conditional independence assumption \cite{dawid1980conditional} and counterfactual reasoning \cite{rafetseder2013counterfactual}. For treatment variable $u_i$, a binary treatment is employed to evaluate the treatment effect of the outcome. The TE of utterance $u_i$ in dialogue $D$ is defined as:
\begin{equation}
    \text{TE}(u_i) = f(D)-f(D\setminus u_i),
\end{equation}
where $D\setminus u_i$ represents the dialogue when utterance $u_i$ is disturbed, $f(\cdot)$ is the perplexity \cite{1995From} of the language model to generate correct responses. To eliminate the potential impact of changes in other co-variables after the utterance absence, we use meaningless utterances of similar lengths, such as `hello,' `thank you,' \emph{etc.}, to replace $u_i$ to construct the counterfactual condition of the absence treatment. 

\begin{figure}[!t]
    \centering
    \begin{minipage}[t]{0.49\columnwidth}
        \centering      \includegraphics[width=1\linewidth]{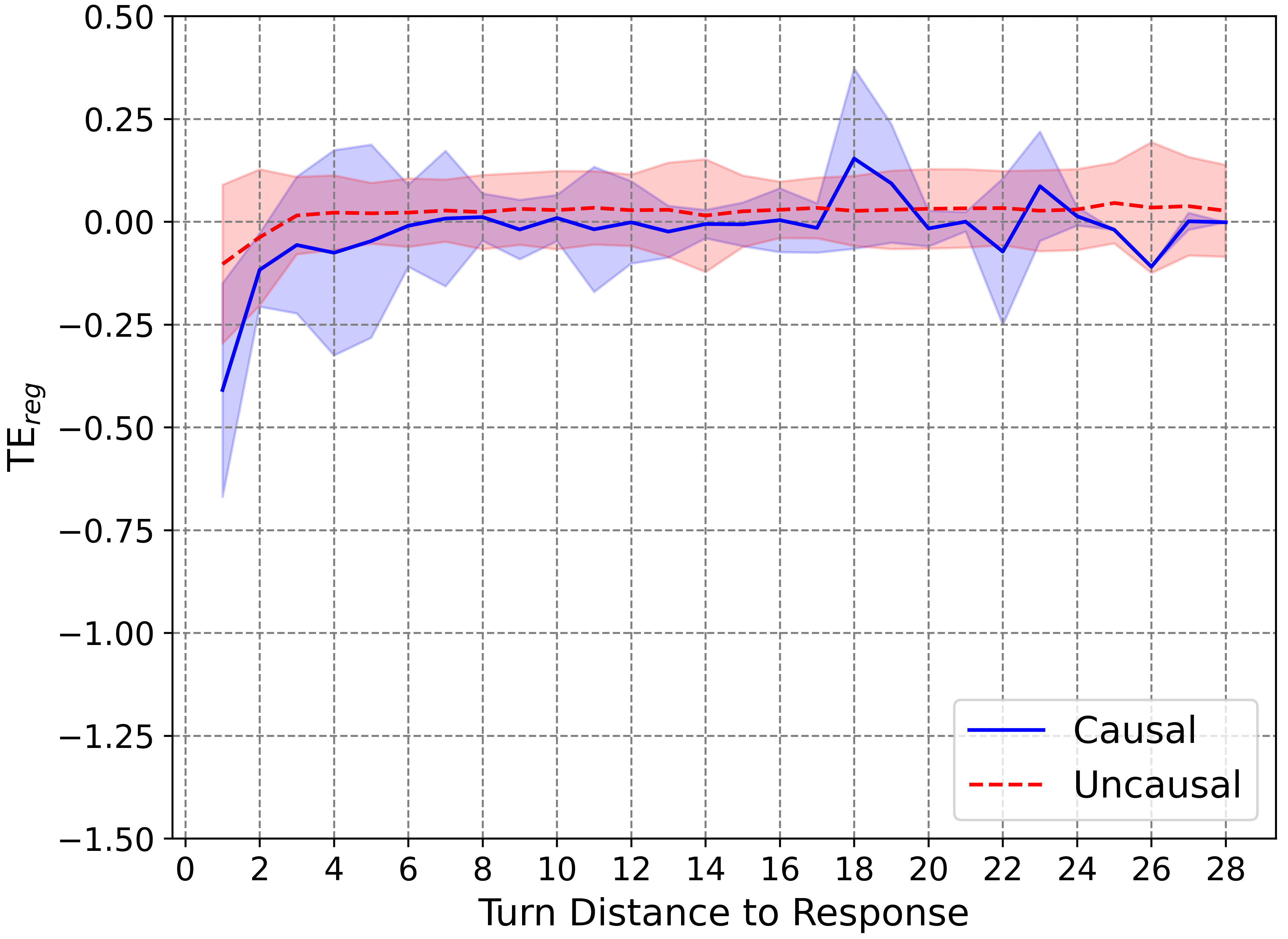}
        \centerline{(a) Raw}
    \end{minipage}
    \begin{minipage}[t]{0.49\columnwidth}
        \centering      \includegraphics[width=1\linewidth]{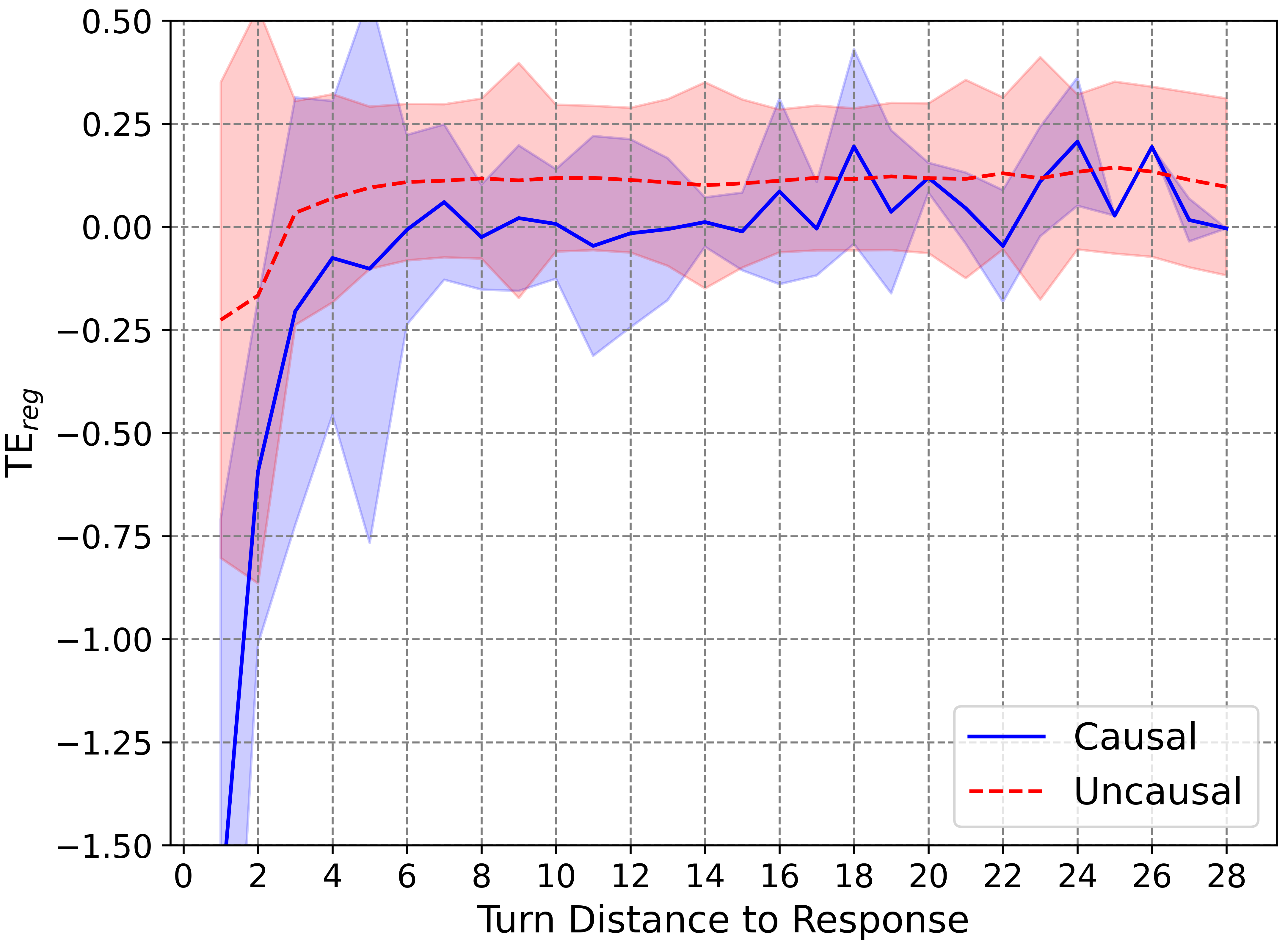}
        \centerline{(b) Finetune}
    \end{minipage}
    \begin{minipage}[t]{0.49\columnwidth}
        \centering      \includegraphics[width=1\linewidth]{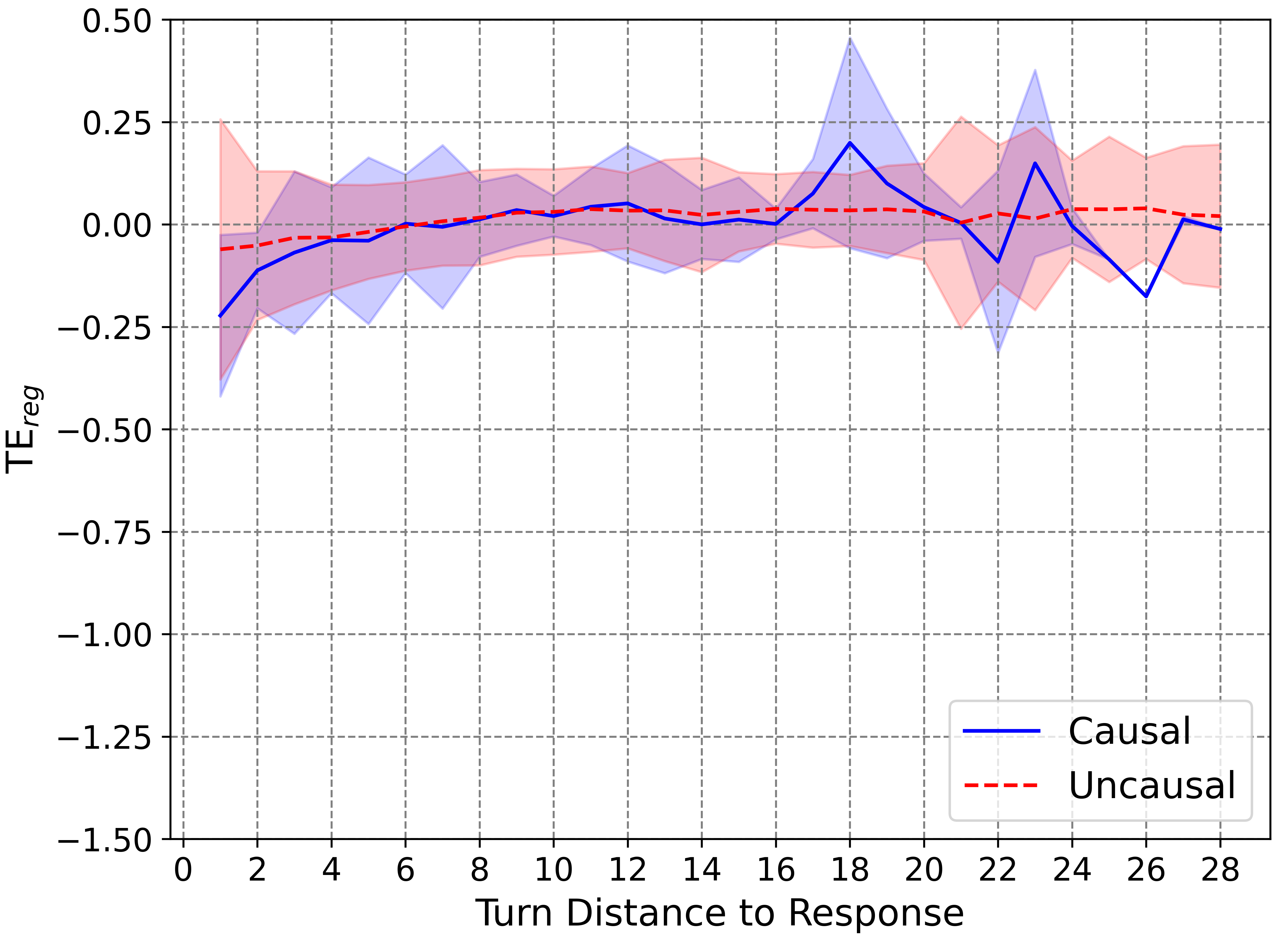}
        \centerline{(c) w/o position embedding}
    \end{minipage}
    \begin{minipage}[t]{0.49\columnwidth}
        \centering      \includegraphics[width=1\linewidth]{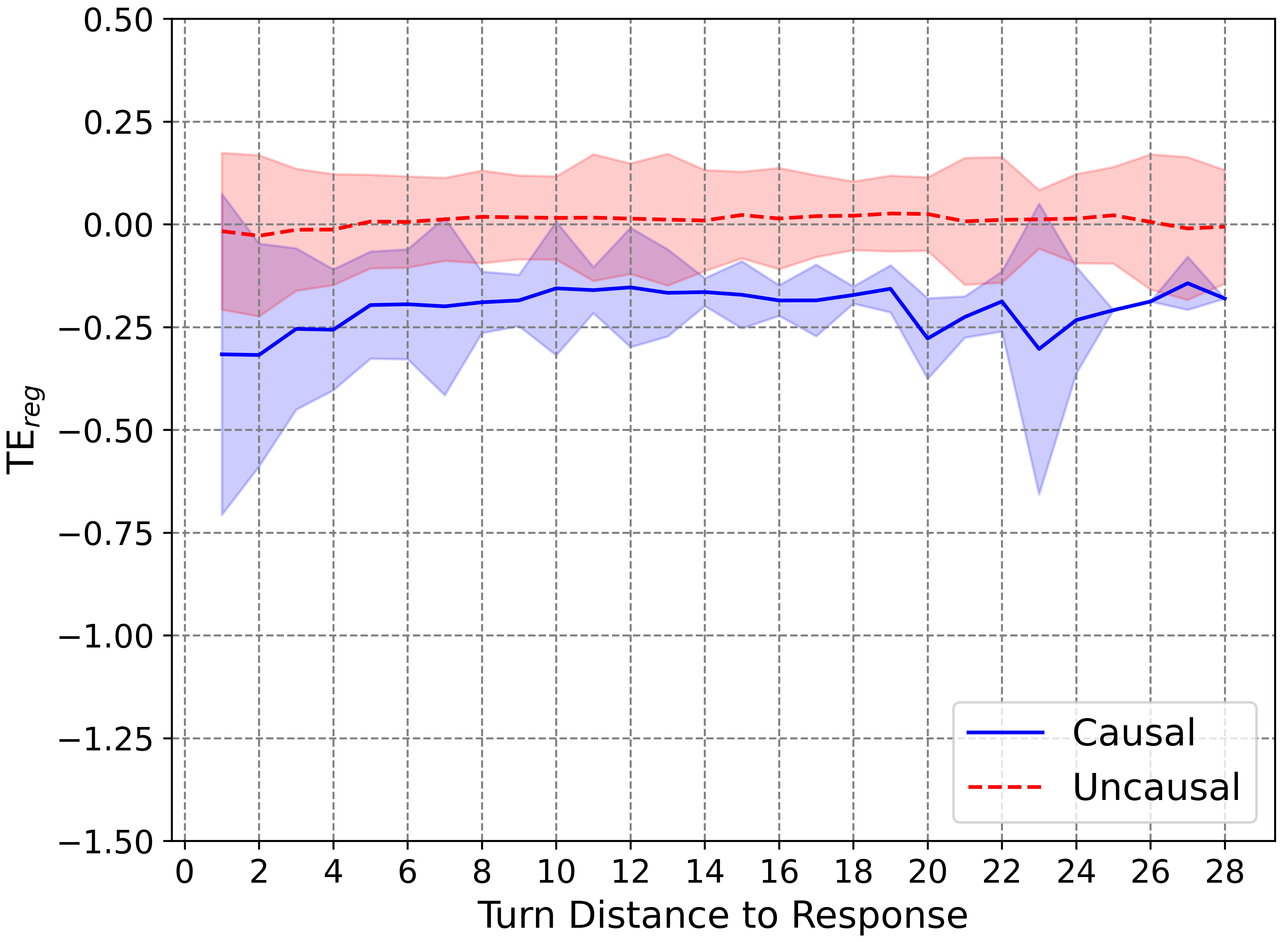}
        \centerline{(d) Ours}
    \end{minipage}
    \caption{Llama2-7B-chat's ability to identify causally relevant utterances in the CGDIALOG dataset.}
    \label{fig:ppl}
\end{figure}

\subsection{Causal Identification and Position Bias}

Outstanding language models should adeptly identify causal correlations in dialogues. Specifically, the TE of causally relevant utterances should be significantly higher than the TE of non-causal utterances. Unfortunately, Abraham \etal \shortcite{abraham2022cebab} observe that language models lack this identification ability. 
We further observe that the model's causal identification ability is strongly tied to the position of causally relevant utterances in the dialogue. The normalized treatment effect $\text{TE}_{\text{reg}}(u_i)$ is employed to measure the causal correlation between utterances and responses. 
\begin{equation}
    \text{TE}_{\text{reg}}(u_i) = [f(D)-f(D\setminus u_i)]/f(D).
\end{equation}

As shown in Figure \ref{fig:ppl} (a) and (b), we find that: (1) The LLMs can only identify the causal correlation in the last 1 and 2 turns of dialogue. (2) Irrespective of whether the utterances are relevant or not, LLMs consistently exhibit greater sensitivity to perturbations in the last few turns of dialogues. (3) Despite fine-tuning on domain data enhancing the model's sensitivity to causally relevant utterances, the accurate distinction remains challenging in longer dialogue histories. Similar results are observed on Qwen-14B-chat~\cite{bai2023qwen} (Figure \ref{app:ppl} in Appendix).

This indicates that LLMs have the potential to distinguish between relevant and irrelevant utterances. One way to eliminate the influence of position information is to remove all position embedding, as shown in Figure \ref{fig:ppl} (c). After removing position embeddings, the model no longer focuses on position-specific utterances. Unfortunately, it also results in the damage of semantic information, rendering the LLMs incapable of identifying relevant utterances.

\begin{figure*}[!t]
    \centering
    \includegraphics[width=1\textwidth]{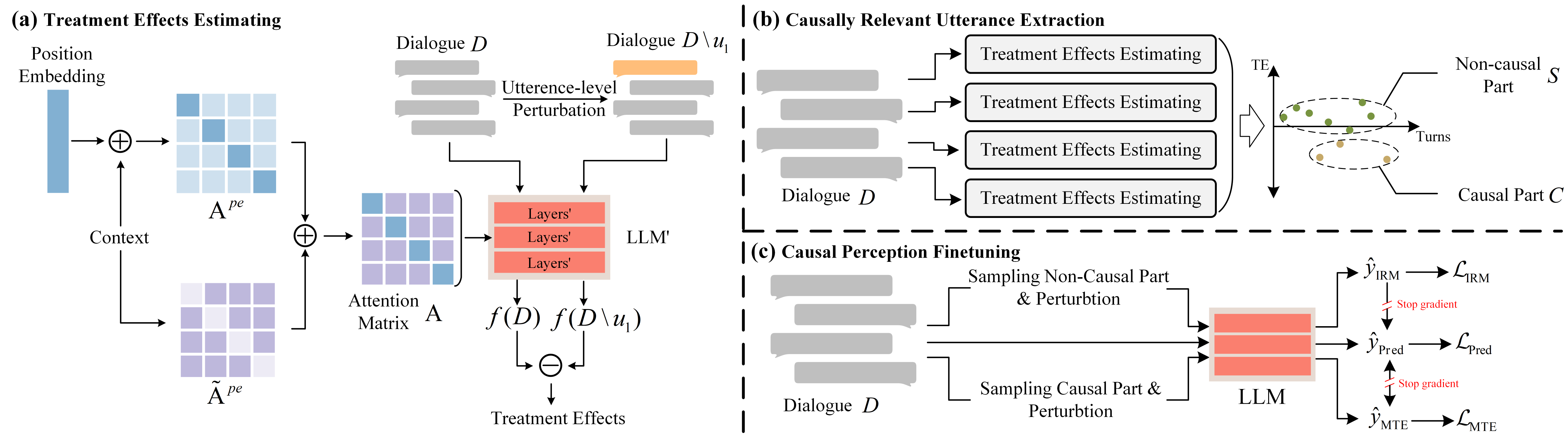}
    \caption{The Framework of our proposed method.}
    \label{fig:model}
\end{figure*}

\subsection{Causally Relevant Utterance Extraction}
To achieve a balance between semantic information and position bias, we propose a sentence-level local-position awareness method for each layer of LLMs. Specifically, we restrict position information within sentences, and inter-sentence attention only uses semantic correlations. To ensure the method's applicability to models employing different positional embeddings, we directly modify the attention matrix. The model utilizes attention with position embeddings $\mathbf{A}_{t,s}^{pe}$ when the input words are in the same sentence. In contrast, when the input words are not in the same sentence, the attention is without position embeddings $\tilde{\mathbf{A}}_{t,s}^{pe}$:
\begin{equation}
    \mathbf{A}_{t,s}=\left\{\begin{array}{l}
    \mathbf{A}_{t,s}^{pe} ,\text { if } x_t,x_s\text{ in the same utterance},\\
    \tilde{\mathbf{A}}_{t,s}^{pe}\times m_A^{pe}/\tilde{m}_A^{pe} ,\text { else } ,
\end{array}\right.
\end{equation}
where $x_t$ and $x_s$ denote the $t$-th and $s$-th input word, $m_A^{pe}$ and $\tilde{m}_A^{pe}$ are the average values of attention matrices $\mathbf{A}^{pe}$ and $\tilde{\mathbf{A}}^{pe}$ respectively, used to balance the difference between two attention matrices.

After fine-tuning, the model's performance is shown in Figure \ref{fig:ppl} (d). It can be observed that our method effectively mitigates position bias, leading to improved identification of causally relevant utterances. To extract the minimal causally relevant utterance set in dialogues, we individually measure the TE of each utterance in the dialogue, denoted as [$\text{TE}(u_1), \text{TE}(u_2), ..., \text{TE}(u_{|D|})$], where $|D|$ represents the number of dialogue turns. Due to the difference in TE between causally relevant and non-causally relevant utterances, we employ a simple clustering algorithm, K-means \cite{krishna1999genetic}, to obtain the causally relevant $C$ and the non-causally relevant utterance set $S$. The initial clustering centers of the two sets are initialized as the minimum and median of the input data, respectively. The experiment in Appendix \ref{app:independent} demonstrates that utterances are conditionally independent and the causally relevant utterance extraction method based on single-sentence perturbation and clustering does not get stuck in locally optimal solutions.

After verifying the effectiveness of our method with the 88.6\% precision on the CGDIALOG test set (refer to Appendix \ref{app:extract}), we extract relevant utterances from two long-term dialogue datasets, ESConv \cite{liu2021towards} and MSC \cite{DBLP:conf/acl/XuSW22}. We calculate the position distribution $Q$ of causally relevant utterances, where $q_i \in Q$ represents the frequency of causally relevant utterances with the $i$-th turn distance to the response. A severe imbalance is observed in the position distribution of causally relevant utterances, which might be the cause of model position bias. For more detailed statistical results, refer to Appendix \ref{app:causal_static}.

\subsection{Causal Perception Finetuning}
The fine-tuning of LLMs follows the paradigm of instruction fine-tuning, where instructions and dialogue $D$ are concatenated and fed into the model for generating responses $R$. 
\begin{equation}
    \begin{aligned}
        p(R) & =p(R \mid instruction, D) \\    
        & =\prod_{t} p\left(r_{t+1} \mid instruction, D, r_{1}, r_{2}, \ldots, r_{t}\right),
    \end{aligned}
\end{equation}

During the fine-tuning process, our objective is for the model to acquire domain knowledge from the data while being sensitive to causal correlations. The loss during the fine-tuning process is divided into two parts: prediction loss and causal perception loss. The prediction loss ensures that the model accurately generates the gold responses and captures domain knowledge in the training data. The causal perception loss is employed to enhance the model's sensitivity to causal correlations.
\begin{equation}
    \mathcal{L} = \mathcal{L}_{\text{Pred}} + \underbrace{\alpha \mathcal{L}_{\text{IRM}} + \beta \mathcal{L}_{\text{MTE}}}_{\text{causal perception}},
\end{equation}
where $\alpha$ and $\beta$ are used to scale the three losses to similar orders of magnitude. The prediction loss is to maximize the cross-entropy between model predictions and gold responses.
\begin{equation}
    \mathcal{L}_{\text{Pred}} = - \sum_{r_t \in R} log(p(r_t \mid instruction, D)).
    \label{eq:raw}
\end{equation}

For the sake of brevity, the \emph{instruction} in the formula is omitted following. The causal perception loss is designed to require the model to focus on the causally relevant utterances while being sensitive to perturbations in environmental variables, such as non-relevant utterances. The causal perception loss consists of two parts: invariant risk minimization $\mathcal{L}_{\text{IRM}}$ and maximizing treatment effect $\mathcal{L}_{\text{MTE}}$.

\noindent \textbf{Invariant risk minimization (IRM)} \cite{arjovsky2019invariant} drives the model to grasp causal invariance across environments, that is, the model's outcomes should remain consistent across various environments, which are constructed by perturbing non-causally relevant utterances. For dialogue $D=\{u_1, u_2, \dots, u_{|D|}\}$ consisting of multiple utterances, we construct counterfactual dialogue $D\setminus u_i$ in which non-causally relevant utterances $u_i \in S$ are replaced. To ensure that the replacements do not significantly alter the semantic structure of dialogues, we randomly select utterances for substitution from non-causally relevant utterances in other dialogues. Invariant risk minimization loss minimizes the KL divergence of responses generated by counterfactual and original dialogues.
\begin{equation}
    \mathcal{L}_{\text{IRM}} = \sum_{r_i \in R} KL(p_{sg}(r_i|D)||p(r_i|D\setminus u_i)), u_i \in S,
    \label{eq:irm}
\end{equation}
where $sg$ means stoping gradient during backpropagation.

\noindent \textbf{Maximizing treatment effect (MTE)} \cite{marshall2005working} aims to help the model learn the consistency relationship between responses and corresponding causal variables. When causally relevant utterances are replaced with those from other dialogues, the loss expects models to generate utterances that exhibit maximum dissimilarity compared to the original dialogue model output.
\begin{equation}
    \mathcal{L}_{\text{MTE}} = - \sum_{r_i \in R} KL(p_{sg}(r_i|D)||p(r_i|D\setminus u_i)), u_i \in C.
    \label{eq:mte}
\end{equation}




\noindent \textbf{Sampling strategy.}  
To overcome the imbalance position distribution of relevant utterances, we employ a sampling strategy that involves enforcing more perturbations at positions with the low likelihood of being a relevant utterance in the dataset. At the dialogue level, multiple perturbations are performed for each dialogue when calculating causal perception loss. The number of times $n$ for both auxiliary tasks IRM and MTE is determined through the following formula:
\begin{equation}
    n = \lfloor |C|/\sum_{u_i\in C}(q_{|D|-i}) \rfloor,
\end{equation}
where $\lfloor \cdot \rfloor$ denotes rounding down, $|\cdot|$ represents the number of elements in the set, and $q_{|D|-i}$ represents the frequency of the $i$-th utterance in the dataset being a causally relevant utterance.

At the utterance level, each time the auxiliary task is calculated, the probability $S(u_i)$ of perturbing each sentence $u_i$ is different. For IRM and MTE, the probability that utterance $u_i$ is perturbed is,
\begin{equation}
    S_{\text{IRM}}(u_i)=\left\{\begin{array}{l}
    0 ,\text { if } u_i \in C,\\
    q_{|D|-i},\text { if } u_i \in S.
\end{array}\right.
\end{equation}

\begin{equation}
\small
    S_{\text{MTE}}(u_i)=\left\{\begin{array}{l}
    (1/q_{|D|-i})/\sum_{u_j \in C}{(1/q_{|D|-j})},\text { if } u_i \in C,\\
    0,\text { if } u_i \in S,
\end{array}\right.
\end{equation}

The process of causal perception finetuning is shown in Algorithm \ref{alg:finetune} in the Appendix.

\section{Experiments}

\subsection{Datasets} 
To evaluate the effectiveness of our proposed method, following previous works~\cite{wang2023recursively,feng2023less}, we conduct experiments on two widely used benchmark datasets, \textbf{ESConv}~\cite{liu2021towards} and \textbf{MSC}~\cite{DBLP:conf/acl/XuSW22}, for long-term dialogue. 
We use the same data preprocessing and train/valid/test splitting strategy as in \cite{feng2023less}. More details of the datasets are shown in the Appendix \ref{app:dataset}.

\subsection{Baselines}
To demonstrate the effectiveness of our proposed method, we compare it with three kinds of baselines:
\textbf{(1) Raw and finetuned LLMs} have outstanding performance in open-domain dialogue tasks. The model can improve its performance in the tasks during fine-tuning in the corresponding task domain.
\textbf{(2) Long-term dialogue methods.} \textbf{RSM}~\cite{wang2023recursively} continuously summarizes long-term dialogues and uses the summary as external memory to alleviate the memory forgetting of long-term dialogues in LLMs. \textbf{CONSTRAIN}~\cite{feng2023less} assumes that except for the last sentence in the dialogues, there is only one sentence relevant to the response. Relevant utterances in dialogue history are retrieved through a trained language model and concatenated with the last sentence as input for response generation. 
\textbf{(3) Position debiasing methods.} \textbf{RPP}~\cite{amor2023technical} is extended to sentence-level position random perturbation to ensure that the training data no longer has imbalances in position distribution. \textbf{ZOE}~\cite{liu2024zero} fits both gold response and the suboptimal response generated by the original model to enforce consistency between the fine-tuned and original model.
For the fairness of the experiments, the backbones in all baselines are replaced by two widely used LLMs, Llama2-7B-chat and Qwen-14B-chat. The Appendix \ref{app:llms} presents details of the two LLMs.

\begin{table*}[!t]
\centering
\resizebox{\textwidth}{!}{
\small
\begin{tabular}{lllcccccccc}
\toprule
\multicolumn{1}{l}{\textbf{Dataset}} & \multicolumn{1}{l}{\textbf{Backbone}} & \textbf{Method} & \textbf{BLEU-1} & \textbf{BLEU-2} & \textbf{ROUGE-L} & \textbf{Distinct-1} & \textbf{Distinct-2} & \textbf{Relevance} & \textbf{Fluency} & \textbf{Informativeness}  \\ \midrule
\multirow{14}{*}{MSC}                & \multirow{7}{*}{Llama2-7B}            & (1) Raw LLM          & 0.0911          & 0.0250          & 0.0860           & 0.0353              & 0.3051              & 1.50               & 1.84             & 1.43                     \\
                                     &                                       & (1) Finetuned LLM    & 0.1037          & 0.0283          & 0.0995           & 0.0371              & 0.3288              & 1.56               & \textbf{1.87}    & 1.49                     \\
                                     &                                       & (2) RSM              & 0.1127          & 0.0336          & 0.1117           & 0.0373              & 0.3150              & 1.62               & 1.82             & 1.64                     \\
                                     &                                       & (2) CONSTRAIN        & \underline{0.1167}    & \underline{0.0342}    & \underline{0.1118}     & 0.0118              & 0.1410              & \underline{1.68}         & \underline{1.86}       & \underline{1.69}               \\
                                     &                                       & (3) RPP              & 0.0927          & 0.0294          & 0.0877           & \underline{0.0473}        & \underline{0.3650}        & 1.47               & \textbf{1.87}    & 1.48                     \\
                                     &                                       & (3) ZOE              & 0.1076          & 0.0328          & 0.1109           & 0.0446              & 0.3416              & 1.61               & 1.84             & 1.65                     \\
                                     &                                       & CPD (Ours)           & \textbf{0.1245} & \textbf{0.0441} & \textbf{0.1214}  & \textbf{0.0655}     & \textbf{0.4233}     & \textbf{1.77}      & \underline{1.86}       & \textbf{1.78}            \\ \cmidrule{2-11}
                                     & \multirow{7}{*}{Qwen-14B}             & (1) Raw LLM          & 0.0973          & 0.0321          & 0.1041           & 0.0443              & 0.3199              & 1.51               & \underline{1.86}       & 1.49                     \\
                                     &                                       & (1) Finetuned LLM    & 0.1199          & 0.0364          & 0.1044           & 0.0486              & 0.3260              & 1.54               & 1.85             & 1.52                     \\
                                     &                                       & (2) RSM              & 0.1175          & 0.0379          & 0.1082           & 0.0526              & 0.3437              & 1.64               & 1.84             & 1.65                     \\
                                     &                                       & (2) CONSTRAIN        & \underline{0.1258}    & \underline{0.0451}    & \underline{0.1233}     & 0.0204              & 0.1474              & \underline{1.73}         & \underline{1.86}       & \underline{1.70}               \\
                                     &                                       & (3) RPP              & 0.1076          & 0.0345          & 0.1059           & \underline{0.0646}        & \underline{0.4416}        & 1.53               & 1.85             & 1.50                     \\
                                     &                                       & (3) ZOE              & 0.1124          & 0.0318          & 0.1100           & 0.0632              & 0.4282              & 1.63               & 1.86             & 1.67                     \\
                                     &                                       & CPD (Ours)           & \textbf{0.1462} & \textbf{0.0519} & \textbf{0.1381}  & \textbf{0.0887}     & \textbf{0.5389}     & \textbf{1.82}      & \textbf{1.87}    & \textbf{1.84}            \\ \midrule
\multirow{14}{*}{ESConv}             & \multirow{7}{*}{Llama2-7B}            & (1) Raw LLM          & 0.0713          & 0.0181          & 0.0739           & 0.0578              & 0.3723              & 1.54               & 1.85             & 1.47                     \\
                                     &                                       & (1) Finetuned LLM    & 0.0842          & 0.0286          & 0.1047           & 0.0614              & 0.4018              & 1.57               & 1.85             & 1.52                     \\
                                     &                                       & (2) RSM              & \underline{0.0949}    & \underline{0.0318}    & \underline{0.1098}     & 0.0766              & 0.4404              & \underline{1.64}         & \textbf{1.87}    & \underline{1.68}               \\
                                     &                                       & (2) CONSTRAIN        & 0.0919          & 0.0276          & 0.1038           & 0.0118              & 0.1293              & 1.63               & 1.84             & 1.67                     \\
                                     &                                       & (3) RPP              & 0.0670          & 0.0258          & 0.0972           & \underline{0.0870}        & \underline{0.4679}        & 1.50               & 1.80             & 1.44                     \\
                                     &                                       & (3) ZOE              & 0.0943          & 0.0297          & 0.1041           & 0.0821              & 0.4532              & 1.61               & 1.85             & 1.67                     \\
                                     &                                       & CPD (Ours)           & \textbf{0.1091} & \textbf{0.0468} & \textbf{0.1324}  & \textbf{0.0976}     & \textbf{0.5094}     & \textbf{1.79}      & \underline{1.86}       & \textbf{1.81}            \\ \cmidrule{2-11}
                                     & \multirow{7}{*}{Qwen-14B}             & (1) Raw LLM          & 0.1077          & 0.0326          & 0.1018           & 0.0744              & 0.4068              & 1.64               & \underline{1.86}             & 1.49                     \\ 
                                     &                                       & (1) Finetuned LLM    & 0.1145          & 0.0372          & 0.1118           & 0.0825              & 0.4230              & 1.66               & 1.85             & 1.53                     \\
                                     &                                       & (2) RSM              & \underline{0.1237}    & \underline{0.0379}    & \underline{0.1168}     & 0.0874              & 0.4413              & \underline{1.74}         & 1.84             & \underline{1.73}               \\
                                     &                                       & (2) CONSTRAIN        & 0.1205          & 0.0301          & 0.1122           & 0.0302              & 0.2564              & 1.73               & \textbf{1.87}    & 1.71                     \\
                                     &                                       & (3) RPP              & 0.1095          & 0.0307          & 0.1056           & \underline{0.0939}        & \underline{0.4782}        & 1.59               & 1.84             & 1.54                     \\
                                     &                                       & (3) ZOE              & 0.1195          & 0.0364          & 0.1147           & 0.0847              & 0.4437              & 1.69               & \underline{1.86}       & 1.70                     \\
                                     &                                       & CPD (Ours)           & \textbf{0.1489} & \textbf{0.0591} & \textbf{0.1442}  & \textbf{0.1125}     & \textbf{0.5459}     & \textbf{1.84}      & \textbf{1.87}    & \textbf{1.88}            \\
\bottomrule
\end{tabular}
}
\caption{The performance of our proposed method and all baselines. The results of the best and the second performance in each column are in \textbf{boldface} and \underline{underlined}, respectively. }
\label{tab:overall_result}
\end{table*}

\begin{table*}[!t]
\centering
\resizebox{\textwidth}{!}{
\small
\begin{tabular}{llccccccccccc}
\toprule
\multirow{2}{*}{\textbf{Backbone}} & \multirow{2}{*}{\textbf{Methods}} & \multicolumn{5}{c}{\textbf{MSC}}                                                                 &  & \multicolumn{5}{c}{\textbf{ESConv}}                                                              \\ \cmidrule{3-7} \cmidrule{9-13} 
                                   &                                   & \textbf{BLEU-1} & \textbf{BLEU-2} & \textbf{ROUGE-L} & \textbf{Distinct-1} & \textbf{Distinct-2} &  & \textbf{BLEU-1} & \textbf{BLEU-2} & \textbf{ROUGE-L} & \textbf{Distinct-1} & \textbf{Distinct-2} \\ \cmidrule{1-7} \cmidrule{9-13} 
\multirow{4}{*}{Llama2-7B}         & CPD                                                  & 0.1245                              & 0.0441                              & 0.1214                               & 0.0655                                  & 0.4233                                  &  & 0.1091                              & 0.0468                              & 0.1324                               & 0.0976                                  & 0.5094                                  \\
                                   & w/o IRM                                               & 0.1036                              & 0.0327                              & 0.1102                               & 0.0523                                  & 0.4063                                  &  & 0.0906                              & 0.0310                              & 0.1116                               & 0.0895                                  & 0.4876                                  \\
                                   & w/o MTE                                               & 0.1165                              & 0.0375                              & 0.1109                               & 0.0423                                  & 0.3420                                  &  & 0.0916                              & 0.0369                              & 0.1260                               & 0.0768                                  & 0.4412                                  \\
                                   & w/o sampling                                          & 0.1074                              & 0.0334                              & 0.1009                               & 0.0570                                  & 0.3882                                  &  & 0.0911                              & 0.0321                              & 0.1194                               & 0.0856                                  & 0.4734                                  \\ \midrule
\multirow{4}{*}{Qwen-14B}          & CPD                                                  & 0.1462                              & 0.0519                              & 0.1381                               & 0.0887                                  & 0.5389                                  &  & 0.1489                              & 0.0591                              & 0.1442                               & 0.1125                                  & 0.5459                                  \\
                                   & w/o IRM                                               & 0.1145                              & 0.0431                              & 0.1128                               & 0.0753                                  & 0.4883                                  &  & 0.1225                              & 0.0466                              & 0.1239                               & 0.1072                                  & 0.5107                                  \\
                                   & w/o MTE                                               & 0.1253                              & 0.0480                              & 0.1271                               & 0.0658                                  & 0.3743                                  &  & 0.1381                              & 0.0534                              & 0.1312                               & 0.0946                                  & 0.4792                                  \\
\multicolumn{1}{l}{}               & w/o sampling                                          & 0.1179                              & 0.0461                              & 0.1257                               & 0.0784                                  & 0.4691                                  &  & 0.1320                              & 0.0496                              & 0.1269                               & 0.1047                                  & 0.5079                                 \\
\bottomrule
\end{tabular}
}
\caption{The performance of ablation study. }
\label{tab:ablation}
\end{table*}

\subsection{Evaluation Metrics}
\noindent \textbf{Automatic Evaluation.} 
(1) \textbf{Word Overlap.} We report \emph{BLEU-n} (n=1, 2)~\cite{DBLP:conf/acl/PapineniRWZ02} and ROUGE-L~\cite{lin2004automatic} to evaluate the coherence and word overlap of generated utterances. 
(2) \textbf{Diversity.} We employ \emph{Distinct-n} (n=1, 2)~\cite{DBLP:conf/naacl/LiGBGD16} to evaluate the diversity of the generated response. 

\noindent \textbf{Human Evaluation.}
We adopt \emph{Relevance}, \emph{Fluency}, and \emph{Informativeness} of the generated utterances with the rating range of [0, 2]. We recruit three experienced annotators to evaluate 100 randomly selected dialogues with a length of more than 20 turns. The Fleiss Kappa is 0.72, indicating consistency in the estimates of annotators. The details are shown in Appendix \ref{app:human}

\begin{table*}[!t]
\centering
\resizebox{\textwidth}{!}{
\small
\begin{tabular}{lp{16cm}}
\toprule
\multirow{11}{*}{Dialogue}               & $\dots$ \\
                                         & Assistant : Hi, do have any issues that you would like to share with me today? \\
                                         & User : \textbf{I am in disputed mod with my friends} \\
                                         & Assistant : I am sorry to hear that. So, you and your friends are not seeing eye to eye? Do you mind telling me what the issue is that you are not agreeing on? \\
                                         & $\dots$ \\
                                         & User : \textbf{They need more money from me, but i couldn't give that much amount} \\
                                         & $\dots$ \\
                                         & User : Sure, it is my new way of relationship to me with others \\
                                         & Assistant : Yes, you're growing and trying to do your best and wanting to expand your boundaries and make them clear. You have taken the right approach and are on the best track to having a successful relationship in the future. Great Job! Thank you for sharing your story with me today. \\ 
                                         & User : \textbf{What kind of people do I dispute with?} \\
                                         & Assistant : \\ \midrule
\multicolumn{1}{l}{\textbf{Qwen-14B-chat}}   & \\ \midrule
\multicolumn{1}{l}{Raw}         & It's important to remember that everyone has different opinions and values which may differ from yours. This type of behavior is very common among toxic relationships where one person takes advantage of another's kindness and generosity without reciprocating. If this continues, it will likely damage both parties involved emotionally and mentally over time. Is there anything else I can assist you with? \\ \midrule
\multicolumn{1}{l}{Finetune}    & It sounds like you may be disputing with those who are taking advantage of you. This type of behavior can happen when someone does not respect our limits and tries to push their own agenda onto another person. If this continues please reach out to local authorities if needed. Is there anything else I can assist you with?\\ \midrule
\multicolumn{1}{l}{RSM}       & \textbf{You have a disputation with someone who has a financial conflication.} It is advisable to cultivate the skill of setting aside conflicts and exploring more rational solutions. I hope to be of assistance to you. \\ \midrule
\multicolumn{1}{l}{CONSTRAIN} & \textbf{You disputed with your friends.} \\ \midrule
\multicolumn{1}{l}{RPP} & That sounds like something which would make anyone worry or be frightened by . Do you have any family around who could support you if worse came to worst? \\ \midrule
\multicolumn{1}{l}{ZOE} & \textbf{It sounds like you may just disagree with your friends'} values and beliefs which causes conflict within your relationships. \\ \midrule
\multicolumn{1}{l}{CPD}      & \textbf{It seems you are in dispute with friends} who have not been responsive to your personal problems and \textbf{have asked for} \textbf{more money than you can provide.} These situations can be challenging, but it’s important to communicate openly and set boundaries in your relationships. Remember, it’s okay to say no when you’re uncomfortable with a request. \\ 
\bottomrule
\end{tabular}
}
\caption{Case study of Qwen-14B-chat. Relevant utterances are marked in boldface.}
\label{tab:case}
\end{table*}

\subsection{Main Result}
The main evaluation results are shown in Tabel~\ref{tab:overall_result}. 

\noindent \textbf{Automatic Evaluation.} CPD attains optimal performance by effectively mitigating the model's position bias and enhancing its causal perception ability. Long-term dialogue methods alleviate the poison of position bias by compressing long-term dialogue history through summarization and retrieval, respectively. Notably, in the ESConv dataset with shorter dialogue turns, summary-based RSM outperforms CONSTRAIN, while retrieval-based CONSTRAIN excels in the longer MSC dataset. These phenomenons underscore that while iterative summarization may retain more information in the short term, it leads to inevitable information loss with multiple iterations. CONSTRAIN neglects the potential significance of multiple causally relevant historical utterances, exhibiting suboptimal performance for shorter dialogues. Additionally, CONSTRAIN performs poorly in diversity indicators, as it overlooks a significant portion of the dialogues' information. Unfortunately, the aforementioned methods, while effectively enhancing LLMs by reducing dialogue history length, fall short of fundamentally improving the models' ability to comprehend extended contexts. Position debiasing methods primarily focus on eliminating position bias but do not address the underlying issue of causal perception in LLMs, resulting in suboptimal performance. RPP, destructing the dialogue structure during fine-tuning, seriously impairs the model's perceptual ability, leading to the poorest performance in word overlap metrics.

\noindent \textbf{Human Evaluation.} Overall, the results of manual and automated assessments align consistently. All methods achieved similar performance in fluency metrics, showcasing the remarkable capabilities of LLMs. The raw and finetuned models underperform in relevance and informativeness metrics due to position bias. CPD enhances both relevance and information content indicators, achieving an average improvement of 5\% and 7\%, respectively, compared to the state-of-the-art baselines. The significant improvement underscores the efficacy of our method in effectively mitigating position bias in LLMs for specific tasks through causal perception finetuning.

\subsection{Ablation Study}
To demonstrate the effectiveness of our proposed method, we establish three variants: (1) w/o IRM: remove the invariant risk minimization loss (Equation \ref{eq:irm}); (2) w/o MTE: remove the maximizing treatment effects loss (Equation \ref{eq:mte}); (3) w/o sampling: remove the positional difference sampling strategy. The experimental results are shown in Table \ref{tab:ablation}.

Experimental results indicate that all ablation variants experienced varying degrees of performance degradation, demonstrating the effectiveness of the components in our method. Furthermore, invariant risk minimization loss can effectively make the model's generation towards gold responses, while maximizing treatment effects loss enhances the diversity of model generation responses. The performance of removing sampling strategy is extremely degraded, proving that it effectively alleviates position imbalance.

\subsection{Case Study}
Table~\ref{tab:case} illustrates an example using Qwen-14B-chat, and the complete dialogue and responses from Llama2-7B-chat can be found in Table \ref{tab:case_app} in the Appendix. The existence of position bias in both the raw and finetuned LLMs leads to generating safe and general but context-irrelevant responses. RSM maintains part of memory capabilities through iterative summarization. CONSTRAIN, despite finding a partially correct answer through retrieval, suffers from substantial information loss in the dialogue, leading to a notable decrease in response diversity. 
RPP disrupts the dialogue structure by altering sentence-level positions during fine-tuning, impairs the model's ability to identify causally relevant utterances, and results in generating context-irrelevant responses. ZOE eliminates part of position bias of LLMs, but the model's causal perception ability is still weak and unable to identify all relevant utterances for response generation. In contrast, CPD benefits from targeted causal perception fine-tuning, effectively perceiving the correct answer and providing informative answers while considering the entire dialogue history.

\section{Conclusion}
In this paper, we analyze the deleterious impact effects of position bias in LLMs on long-term dialogue tasks from a causal perspective. To solve the problem, we propose a model-free Causal Perception long-term Dialogue framework (CPD). 
We extract causally relevant utterances and mitigate position bias through causally perturbed fine-tuning. Specifically, we propose local-position awareness by localizing position information within utterances and further combining it with a perturbation-based method to extract causally relevant utterances. We also propose a causal perception fine-tuning strategy that guides models to focus on causal invariant variables by differently perturbing causally relevant and non-causally relevant utterances in dialogues. A positional difference sampling strategy is employed to address positional imbalances in datasets while maintaining the temporal structure of dialogues. To the best of our knowledge, this is the first work to systematically analyze and alleviate position bias in LLMs from a causal perspective. Experiment results demonstrate the effectiveness of our method in alleviating position bias, resulting in informative and human-like response generation.

\section*{Acknowledgments}
This work was supported in part by the National Natural Science Foundation of China under Grant No. 62276110, No. 62172039 and in part by the fund of Joint Laboratory of HUST and Pingan Property \& Casualty Research (HPL). The authors would also like to thank the anonymous reviewers for their comments on improving the quality of this paper.

\bibliographystyle{named}
\bibliography{ijcai24}

\clearpage
\appendix

\section{Experimental Details}

\begin{table}[t]
\centering 
\small
    \begin{tabular}{lrrr}
    \toprule
    \makebox[0.2\linewidth][l]{\textbf{Dataset}}     & \makebox[0.2\linewidth][r]{\textbf{Dialogue}} & \makebox[0.2\linewidth][r]{\textbf{Utterance}}  & \makebox[0.2\linewidth][r]{\textbf{Average Turns}}  \\ \midrule
    ESConv               & 1,300             & 38,365               & 29.8 \\
    MSC                  & 13,005            & 486,004              & 53.3 \\ \bottomrule                             
    \end{tabular}
\caption{Statistics of the Datasets}
\label{tabel:dataset}
\end{table}
\subsection{Datasets} \label{app:dataset}
Table \ref{tabel:dataset} presents the statistics of ESConv and MSC datasets. 

\begin{itemize}
\item \textbf{ESConv}\footnote{https://github.com/thu-coai/Emotional-Support-Conversation}~\cite{liu2021towards} is a long-term dialogue dataset in the emotion supported domain, composed of 1,300 two-party dialogues between a seeker and an emotion supporter. 
The model will act as a psychologist and provide users with services such as social interactions, mental health support, and customer service chats. 

\item \textbf{MSC}\footnote{https://parl.ai/projects/msc/}~\cite{DBLP:conf/acl/XuSW22} is a long-term multi-session open-domain dialogue dataset. Two humans have a 5-session conversation around their given personalities. In the experiment, we do not enter personality information, only the conversation text.
\end{itemize}

\subsection{Backbones} \label{app:llms}
\begin{itemize}
    \item \textbf{Llama2-7B-chat}\footnote{https://huggingface.co/meta-llama/Llama-2-7b-chat-hf} uses a similar method to Chatgpt for pre-training, while using supervised fine-tuning (SFT) and reinforcement learning with human feedback (RLHF) to iteratively optimize dialogue scenarios. The context window length during training is 4k.
    \item \textbf{Qwen-14B-chat}\footnote{https://huggingface.co/Qwen/Qwen-14B-Chat} uses training fine-tuning technology similar to llama2 to perform targeted optimization of dialogue scenarios. The context window length during model training is 2k, and the inference length is extended to 8k through dynamic NTK-aware interpolation, LogN-Scaling, and window attention.
\end{itemize}

\subsection{Implementation Details}
All of the baselines and our method are implemented in PyTorch. Throughout the experiments, we use Adam optimizer \cite{DBLP:journals/corr/KingmaB14} with 3e-4 initial learning rate and the 128 batch size. All methods are trained for up to 12 epochs. To improve experimental efficiency, we use lora~\cite{hu2021lora} with rank 32 to fine-tune large language models. Both training and inference use 4-bit weight quantization by bitsandbytes~\cite{dettmers2022gpt3}. The optimization solutions we use are widely used and have been verified to have almost no performance degradation compared with full-precision and all-parameters fine-tuning.

\subsection{Human Evaluation Details} \label{app:human}
\emph{Relevance} is used to evaluate the relevance of selected topics and generated sentences to historical conversations and the user's personality. \emph{Fluency} is used to measure the fluency of generated utterances. \emph{Informativeness} is used to evaluate whether the generated utterance revolves around the topics and user personas. The evaluation details are shown in the Tabel \ref{table:human}.

\begin{table}[!t]
\centering
    \begin{tabular}{p{7.5cm}}
    \toprule
    \textbf{Relevance} \\
    2: Fits the user's personality and is related to the current conversation\\
    1: Relevant to one of the user's personality or current conversation, irrelevant or conflicting with the other  \\
    0: Does not resemble any user's personality and context history \\ 
    \midrule
    \textbf{Fluency} \\
    2: Fluent and easy to read \\ 
    1: Grammatically formed \\  
    0: Not a complete sentence or hard to read \\  
    \midrule
    \textbf{Informativeness} \\ 
    2: Have clear and specific meaning \\
    1: Contain a few informative words \\ 
    0: Meaningless sentence  \\
    \bottomrule
    \end{tabular}
\caption{Criteria of human evaluation.}
\label{table:human}
\end{table}

\section{More Experiments}

\begin{figure}[!t]
  \centering
  \includegraphics[width=\columnwidth]{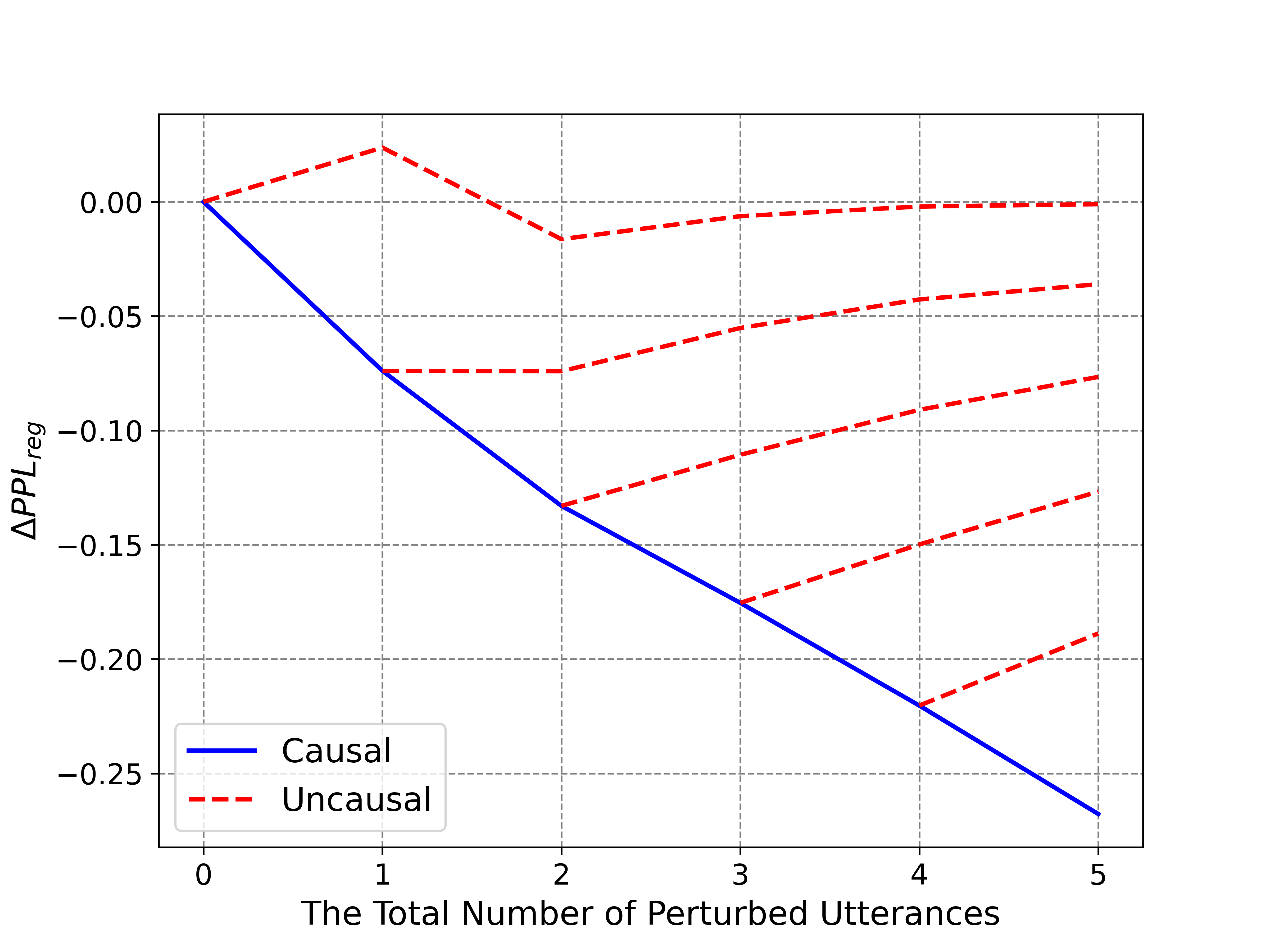}
  \caption{Our method (with Llama2-7B-chat as backbone)’s ability to identify multi causally relevant utterances. The solid blue line depicts the treatment effect as the number of perturbed causally relevant utterances is equal to the x-axis value. The red dashed line represents the treatment effect when the number of perturbed causally relevant utterances matches the x-axis value at the starting point of the blue line and the remains are uncausally relevant utterances.}
  \label{fig:independent}
\end{figure}

\subsection{Conditional Independent Verification} \label{app:independent}
To verify the feasibility of conditionally independent extraction of relevant utterances in dialogues and avoid the method being poisoned by local optimality, we conducted a multi-sentence perturbation test on CGDIALOG dataset. For dialogues with more than 5 relevant utterances in the data set, we randomly select $n$ relevant utterances (blue line) for perturbation, $n \in [0,5]$, and further perturb additional relevant utterances (blue line) or irrelevant utterances (red line). The average PE of the model is shown in Figure~\ref{fig:independent}.

We found that for any set of perturbation-relevant utterances, additional perturbation of relevant utterances always makes PE decrease, and conversely, additional perturbation of irrelevant utterances always makes PE unchanged or increased. The clear difference between the two justifies conditional independence.

Benefiting from the conditional independence between sentences in our method, the time complexity of relevant utterance extraction is O ($n$), where $n$ is the number of utterances in the dialogue. Due to a large number of identical subsequences during the extraction process, our method can be further accelerated by key-value sharing methods such as PagedAttention.

\begin{table}[!t]
\centering
\small
\begin{tabular}{lrrr}
\toprule
    \makebox[0.2\linewidth][l]{\textbf{Method}} & \makebox[0.2\linewidth][r]{\textbf{Precision}} & \makebox[0.2\linewidth][r]{\textbf{Recall}} & \makebox[0.2\linewidth][r]{\textbf{F1-Score}}     \\ \midrule
    Raw                             & 0.5225    & 0.4393 & 0.4773 \\
    Finetune                        & 0.6047    & 0.4756 & 0.5324 \\
    Ours                            & 0.8860    & 0.7852 & 0.8326 \\ \bottomrule
\end{tabular}
\caption{The performance of Llama2-7B-chat in identifying relevant utterances in the CGDIALOG dataset.}
\vspace{-0.5cm}
\label{tab:f1}
\end{table}

\subsection{Causally Relevant Utterance Extraction} \label{app:extract}
Table \ref{tab:f1} shows the performance of Llama2-7B-chat in identifying relevant utterances. The results indicate that local-position awareness can effectively alleviate the impact of position bias on relevant utterance identification and achieve significant performance improvements. The performance of both the raw and finetuned models primarily comes from the selection of the last utterance in the dialogue. 

\begin{figure}[!t]
    \centering
    \begin{minipage}[t]{0.49\columnwidth}
        \centering      \includegraphics[width=1\linewidth]{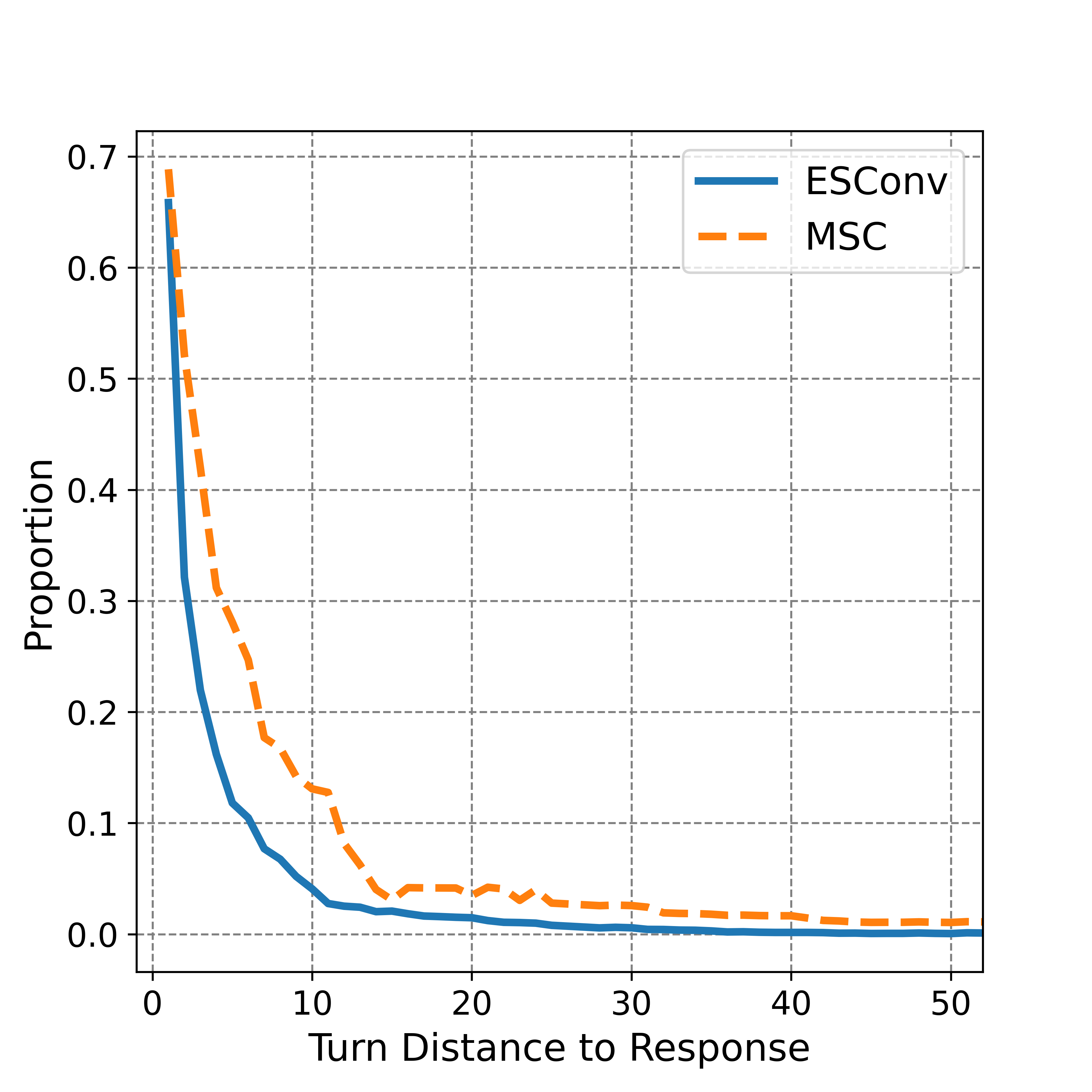}
    \end{minipage}
    \begin{minipage}[t]{0.49\columnwidth}
        \centering      \includegraphics[width=1\linewidth]{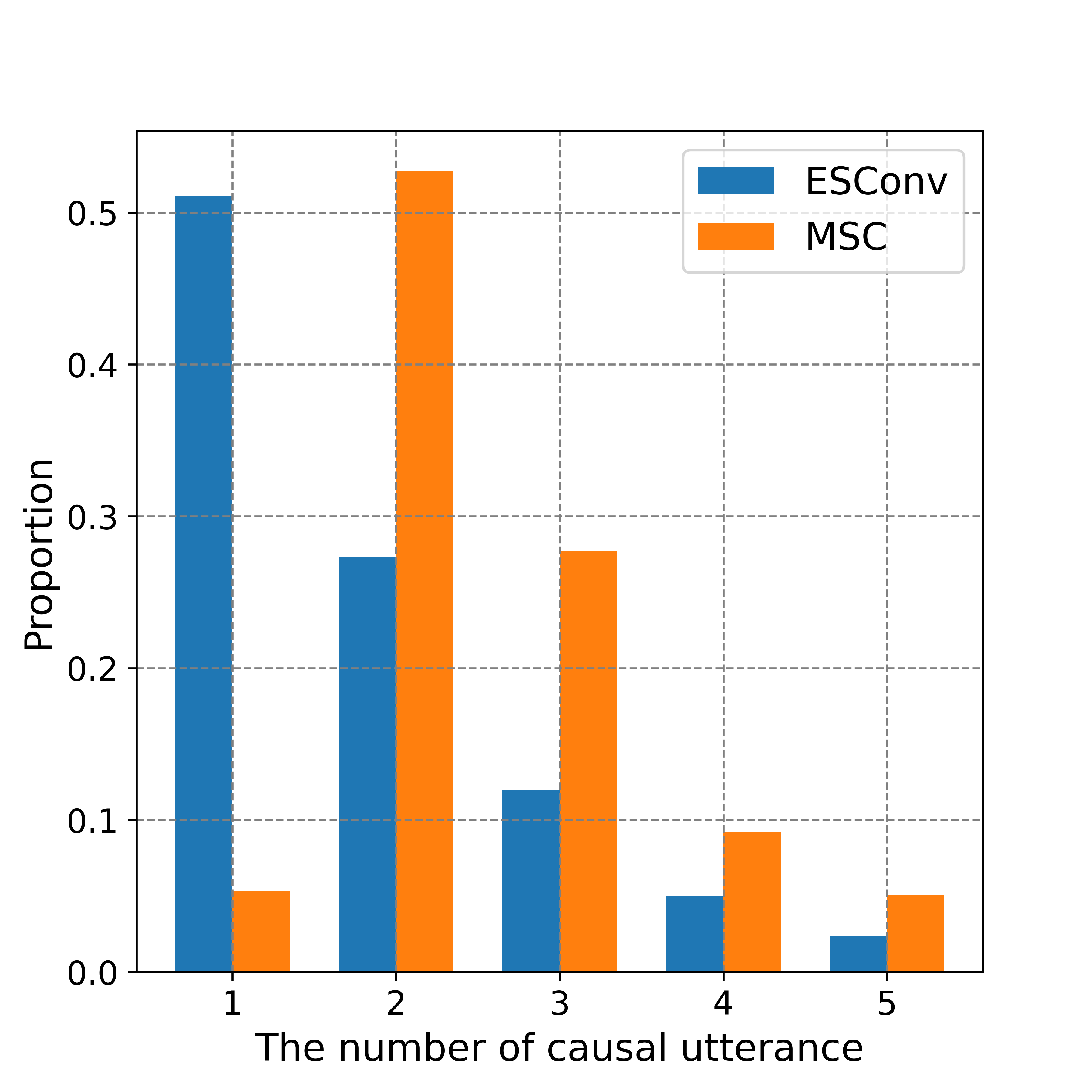}
    \end{minipage}
    \caption{Statistical results of causal correlation dialogues in MSC and ESConv datasets.}
    \label{fig:causal_static}
\end{figure}

\subsection{Causal Relevant Utterance Statistical Results} \label{app:causal_static}
We count the position distribution and number distribution of causally relevant utterances in the two datasets, MSC and ESConv. The statistical results are shown in Figure \ref{fig:causal_static}. We find that the distribution of causally relevant utterances in dialogues has position distribution imbalance, that is, the relevant utterances are mainly distributed in turns close to responses, which is consistent with our intuition. This results in the model being likely to fit position information rather than true causal correlation during both the pre-training and fine-tuning processes. At the same time, we find that the number of relevant utterances and the interval between responses are increasing in MSC, which has more dialogue turns, compared with ESConv. This illustrates the need to enhance the model’s ability to perceive causal correlations and get rid of inherent positional bias in long-term dialogues.

\begin{figure}[!t]
    \centering
    \begin{minipage}[t]{0.49\columnwidth}
        \centering      \includegraphics[width=1\linewidth]{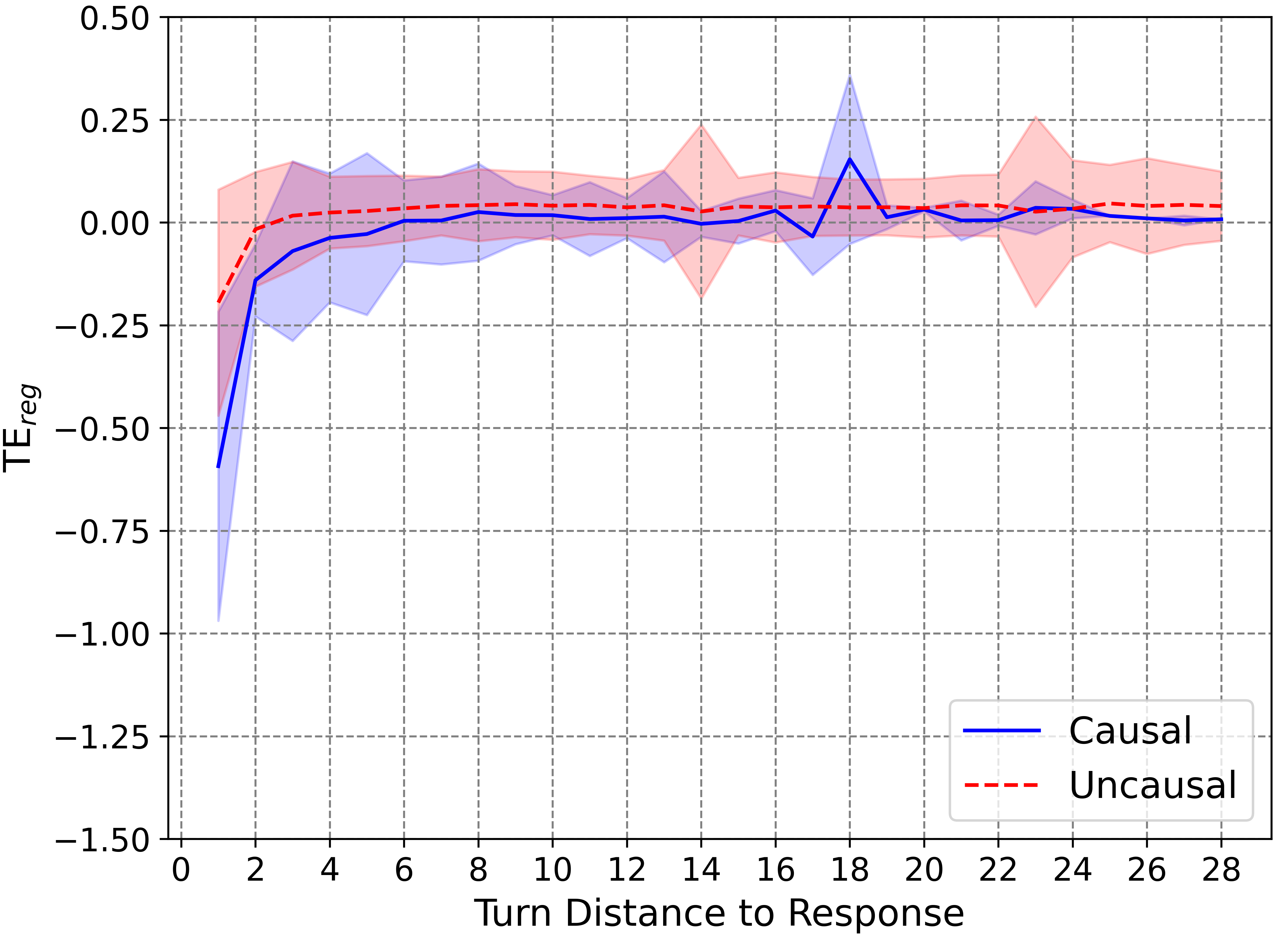}
        \centerline{(a) Raw}
    \end{minipage}
    \begin{minipage}[t]{0.49\columnwidth}
        \centering      \includegraphics[width=1\linewidth]{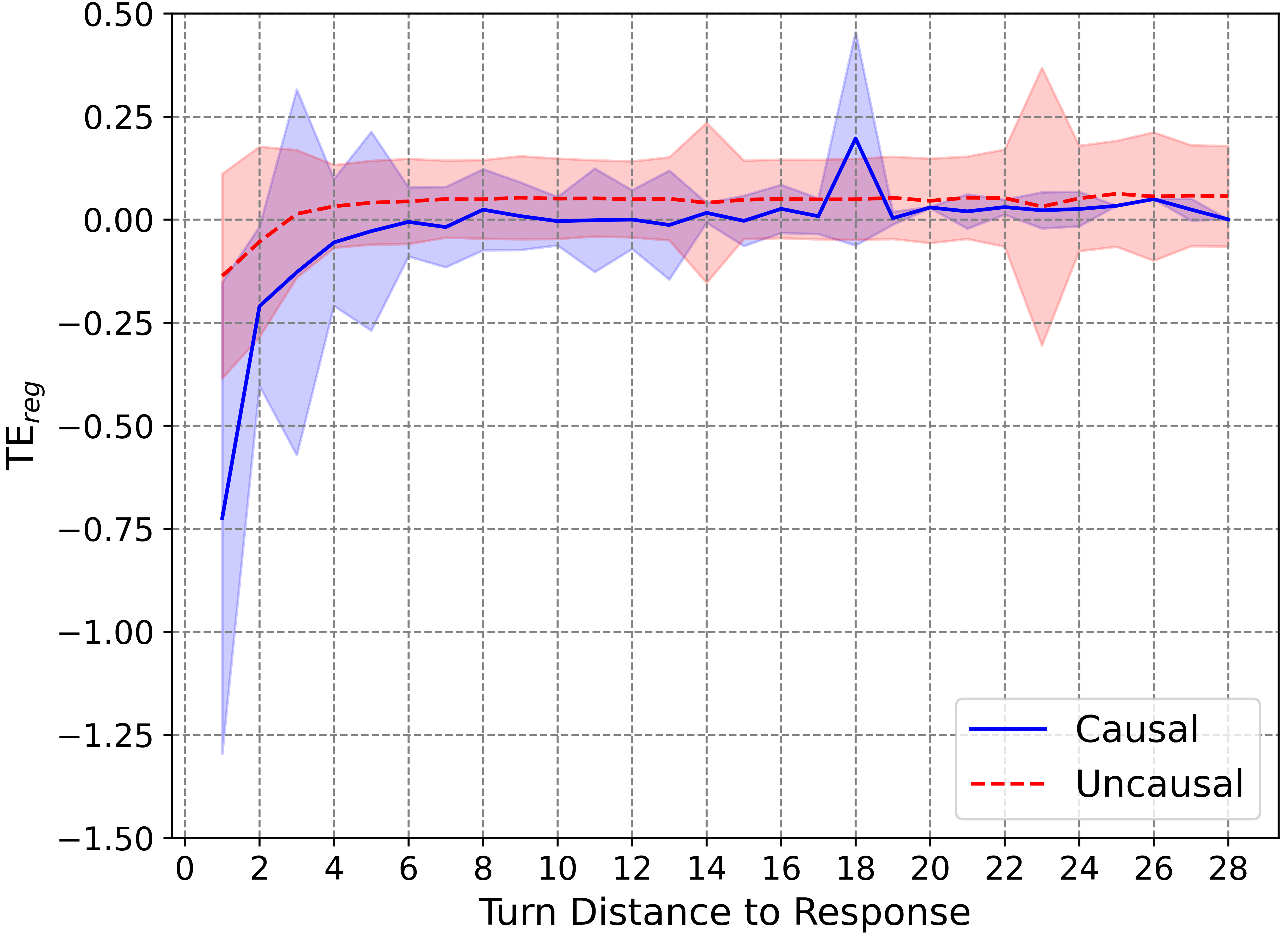}
        \centerline{(b) Finetune}
    \end{minipage}
    \begin{minipage}[t]{0.49\columnwidth}
        \centering      \includegraphics[width=1\linewidth]{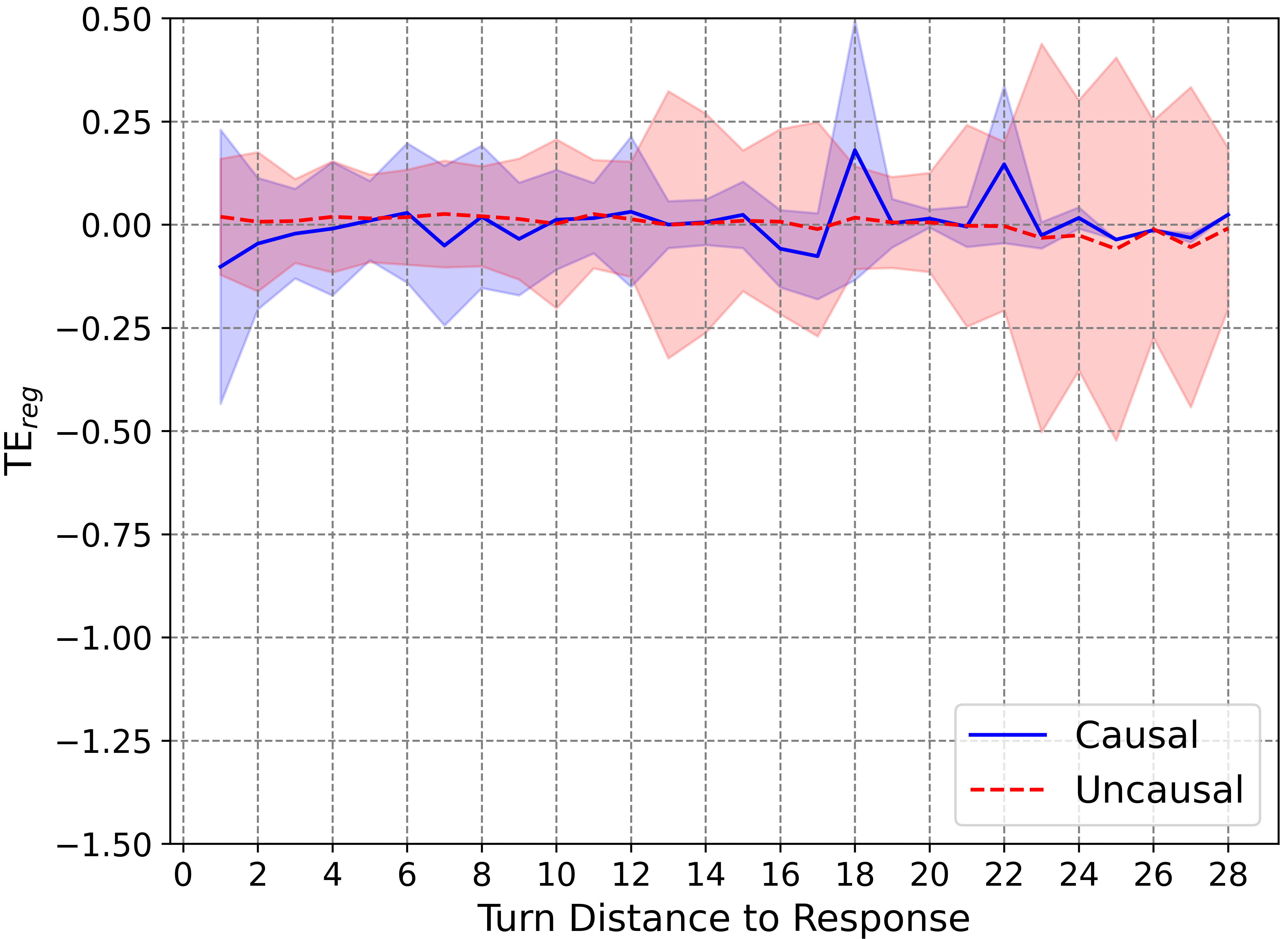}
        \centerline{(c) w/o position embedding}
    \end{minipage}
    \begin{minipage}[t]{0.49\columnwidth}
        \centering      \includegraphics[width=1\linewidth]{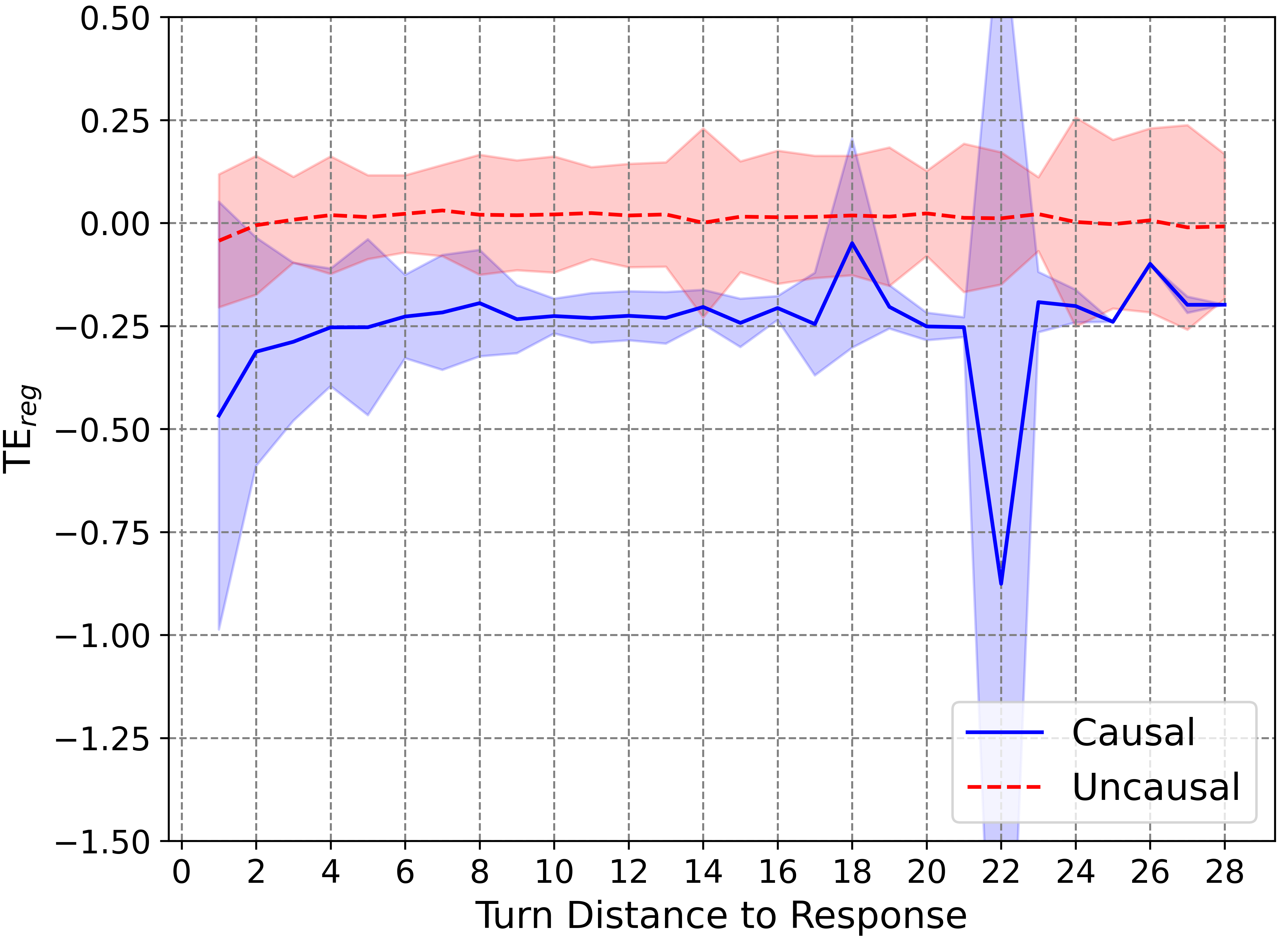}
        \centerline{(d) Ours}
    \end{minipage}
    \caption{Qwen-14B-chat's ability to identify causally relevant utterances.}
    \label{app:ppl}
\end{figure}

\begin{algorithm*}[!t] \small
\caption{Causally Relevant Utterance Extraction}\label{alg:extract}
    \KwData{Dialogue $D=\{u_1, u_2, \dots, u_n\}$}
    Create an empty list named PEs \\
    \ForEach{$u_i$ in $D$}{
        Compute $\text{TE}(u_i)$ as $f(D) - f(D\setminus u_i)$ \\
        Append $\text{TE}(u_i)$ to the list TEs 
    }
    Determine $s_1$, $s_2$ as the minimum and the median of TEs \\
    Initialize K-means cluster centers with $s_1$, $s_2$ \\
    set$_1$, set$_2$ = K-means(TEs) \tcp{return corresponding utterances} 
    \textbf{return} set$_1$ \\
\end{algorithm*}

\begin{algorithm*}[!t] \small
    \caption{Causal perception Finetuning}\label{alg:finetune}
    \KwData{Dataset $\mathcal{D}$, each dialogue $D \in \mathcal{D}$ can be divided into causally relevant $C$ and non-causally relevant utterance set $S$, the set of all utterances and non-causally relevant utterances in the dataset are denoted as $\mathcal{U}$ and $\mathcal{\overline{U}}$, the frequency of causally relevant utterances $Q$.}
    \ForEach{$D \in \mathcal{D}$}{
        Calculate $\mathcal{L}_{\text{Pred}}$ for $D$  \tcp{Equation \ref{eq:raw}}
        $\mathcal{L}$ = $\mathcal{L}_{\text{Pred}}$  \\
       Fix the sampling number $n$ as $\lfloor |C|/\sum_{u_i\in C}(q_{|D|-i}) \rfloor$  \\
        \tcc{Calculate invariant risk minimization loss}
        \For{idx in range n}{
            Sampling a subset $u$ from $S$ such that the sampling probability of $u_i$ is $q_{|D|-i}$ \\
            \ForEach{$u_i \in u$}{
                Substitute $u_i$ in the $D$ witha random utterance in $\mathcal{\overline{U}}$ \\
            }
            Calculate $\mathcal{L}_{\text{IRM}}$ for the dialogue after replacing utterances \tcp{Equation \ref{eq:irm}}
            $\mathcal{L}$ += $\alpha\mathcal{L}_{\text{IRM}}$ \\
        }
        \tcc{Calculate maximizing treatment effects loss}
        \For{$idx$ in range $n$}{
            Sampling a subset $u$ from $C$ such that the sampling probability of $u_i$ is $(1/q_{|D|-i})/\sum_{u_j \in C}{(1/q_{|D|-j})}$ \\
            \ForEach{$u_i \in u$}{
                Substitute $u_i$ in the $C$ with a random utterance in $\mathcal{U}$
            }
            Calculate $\mathcal{L}_{\text{MTE}}$ for the dialogue after replacing utterances \tcp{Equation \ref{eq:mte}}
            $\mathcal{L}$ += $\beta\mathcal{L}_{\text{MTE}}$  
        }
        Calculate the gradient of the loss $\mathcal{L}$ and optimize \\
    }
\end{algorithm*}

\begin{table*}[!t]
\centering
\small
\begin{tabular}{lp{14.8cm}}
\toprule
\multicolumn{1}{l}{Instruct}             & The following is a conversation between an   AI assistant called Assistant and a human user called User. The assistant needs to generate response **briefly** and **precisely** according to the given dialog history. \\ \midrule
\multirow{22}{*}{Dialogue}               & Assistant : Hi, can i help today? Please, tell me about yourself? \\
                                         & User : hi \\
                                         & User : i'm nereida \\
                                         & Assistant : Hi, do have any issues that you would like to share with me today? \\
                                         & User : \textbf{I am in disputed mod with my friends} \\
                                         & Assistant : I am sorry to hear that. So, you and your friends are not seeing eye to eye? Do you mind telling me what the issue is that you are not agreeing on? \\
                                         & User : I explained my personal problems but they are not giving proper response \\
                                         & Assistant : That can be difficult and I can relate to that too. When friends do not respond the way that we feel that they should it can be really hard. Do you have any family that you are close with? \\
                                         & User : \textbf{They need more money from me, but i couldn't give that much amount} \\   & Assistant : I understand the issue of money can cause a lot of problems in a relationship. I had a best friend borrow money once and it lead to her never paying it back and we stopped being friends. Do you plan on trying to talk to your friends soon? Maybe I can give you some advice on how to approach them.\\ 
                                         & User : I want to speak with them because they are in good nature \\
                                         & Assistant : Maybe you could chose a place to meet that is neutral, somewhere at a park or at a cafe. You can ask them to reflect on the good memories in your friendship and then express how you value them as people and friends. Does that sound like something you can try? \\
                                         & User : Yes, i want to do some moderate friendship with them \\
                                         & Assistant : That is a good idea too. Having boundaries is the healthiest part of a relationship so that neither party gets taken advantage of. Setting up lines not to cross is a great idea. \\
                                         & User : I want to do some help to them but i don't do expect more from them \\
                                         & Assistant : I understand that friends can disappoint us sometimes and friendships can take a lot of effort. I would recommend starting a journal. A daily journal can help you express your feelings on paper and it can help you from getting to frustrated and keeping your feelings inside. \\
                                         & User : Yes, some disappointments show us about real face of friends, so it was an experience to me. \\
                                         & Assistant : I am proud of you today because you reached out and have tried to find help outside of yourself. You are growing and trying to do your best. \\
                                         & User : Sure, it is my new way of relationship to me with others \\
                                         & Assistant : Yes, you're growing and trying to do your best and wanting to expand your boundaries and make them clear. You have taken the right approach and are on the best track to having a successful relationship in the future. Great Job! Thank you for sharing your story with me today. \\                    
                                         & User : \textbf{What kind of people do I dispute with?} \\
                                         & Assistant : \\ \midrule
\multicolumn{1}{l}{\textbf{Llama2-7B-chat}}   & \\ \midrule
\multicolumn{1}{l}{Raw}                 & People who care deeply about their relationships often struggle when those bonds become strained due to misunderstandings or unmet emotional demands. This type of behavior is very common among toxic relationships where one person takes advantage of another's kindness and generosity without reciprocating. If this continues, it will likely damage both parties involved emotionally and mentally over time. Is there anything else I can assist you with?  \\ \midrule
\multicolumn{1}{l}{Finetune}             & People who are not open to hearing another persons point of view. Sometimes its best just to walk away from someone if they choose a certain lifestyle or behavior that does not work for you. \\ \midrule
\multicolumn{1}{l}{RSM}       & It sounds like \textbf{you may be disputing with those who have financial problems} or agree with how far you want to take your relationships. Is that correct? \\ \midrule
\multicolumn{1}{l}{CONSTRAIN} & People who will take advanage of you and not respect the line that you set up. \\ \midrule
\multicolumn{1}{l}{RPP} & Are there any other concerns or issues that you would like to share with me today? \\ \midrule
\multicolumn{1}{l}{ZOE} & \textbf{Someone who take your money.} It takes time and patience which both parties must commit themselves fully into if there is hope for reconciliation. \\ \midrule
\multicolumn{1}{l}{Ours}      & \textbf{It seems you're in a dispute with your friends over financial matters}, try to communicate openly, understand their viewpoint, and find common ground for resolution. \\ 
\bottomrule
\end{tabular}
\caption{Case study of Llama2-7B-chat. Relevant utterances are marked in boldface.}
\label{tab:case_app}
\end{table*}

\end{document}